\documentclass{article}

\PassOptionsToPackage{comma,numbers}{natbib}
\usepackage[main, final]{neurips_2026}

\usepackage[utf8]{inputenc} % allow utf-8 input
\usepackage[T1]{fontenc}    % use 8-bit T1 fonts
\usepackage[backref = page, colorlinks = True, linkcolor = violet!70!white, citecolor = violet!70!white]{hyperref}       % hyperlinks
\usepackage{url}            % simple URL typesetting
\usepackage{booktabs}       % professional-quality tables
\usepackage{amsfonts}       % blackboard math symbols
\usepackage{nicefrac}       % compact symbols for 1/2, etc.
\usepackage{microtype}      % microtypography
\usepackage{xcolor}         % colors
\usepackage{amsmath}
\usepackage{mathtools}
\usepackage{algorithm}
\usepackage{algorithmic}
\usepackage{xspace}
\usepackage{wrapfig}
\usepackage{cleveref}
\usepackage{subcaption}
\usepackage{enumitem}     % compact lists (Sec. 3.4 losses)

\usepackage{makecell}    % makecell for multi-line headers
\usepackage{graphicx}    % for resizing

\newcommand{\method}{\mbox{\textsc{Imbue}}\xspace}

\title{Amortized Bayesian Experimental Design with In-Context Knowledge Conditioning}

\author{%
  \textbf{Zhanghu Zhao}$^{1,2}$ \quad
  \textbf{Daolang Huang}$^{1,2}$ \quad
  \textbf{Xinyi Wen}$^{1,2,3}$ \quad
  \textbf{Samuel Kaski}$^{1,2,4}$ \\[0.6em]
  $^1$\,ELLIS Institute Finland \\
  $^2$\,Department of Computer Science, Aalto University, Finland \\
  $^3$\,Department of Computer Science, University of Helsinki, Finland \\
  $^4$\,Department of Computer Science, University of Manchester, UK \\
  [0.6em]
  \texttt{\{zhanghu.zhao, daolang.huang, xinyi.wen, samuel.kaski\}@aalto.fi} \\
}

\begin{document}

\maketitle

\begin{abstract}
Amortized Bayesian experimental design (BED) enables real-time design strategies by shifting acquisition cost offline, but existing methods remain tied to the prior and task distribution they were trained on and cannot exploit additional information available at deployment. We introduce in-context amortized BED with unified knowledge encoding (\method), a BED framework that admits external knowledge at deployment through a unified in-context interface. \method accepts two forms of such knowledge without retraining, namely user-specified prior knowledge and observations from previous instances of the experiment. Both are encoded as additional input tokens and processed in-context with the accumulated observations by a shared Transformer, and a learned reliability filter scores each external token against the observations and excludes inconsistent ones. Across four benchmarks, \method accelerates early-stage information acquisition when the external knowledge is reliable, and remains close to the no-external baseline when it is not.
\end{abstract}

\section{Introduction}
\label{sec:introduction}

A key challenge across disciplines such as psychology \citep{myung2013tutorial, vincent2017darc, watson2017quest+}, physics \citep{dushenko2020sequential}, and drug discovery \citep{lyu2019ultra, masood2024deep} is the design of experiments to infer the unobservable properties of the systems under study. Bayesian Experimental Design (BED) \citep{rainforth2024modern, ryan2016review, chaloner1995bayesian, lindley1956measure} provides a principled framework for this challenge, optimizing the design by maximizing the expected information gain (EIG) about the parameters of interest. Because the traditional adaptive BED loop is computationally demanding and inherently myopic \citep{rainforth2018nesting, foster2019variational}, recent work has turned to amortized policy-based BED \citep{foster2021deep, ivanova2021implicit, blau2022optimizing, hedman2025stepdad, huang2025aline, bracher2025jadai}, which trains a design policy $\pi_\psi$ offline and at deployment maps the accumulated observations directly to the next design in a single forward pass.

\begin{figure}
  \centering
  \includegraphics[width=\textwidth]{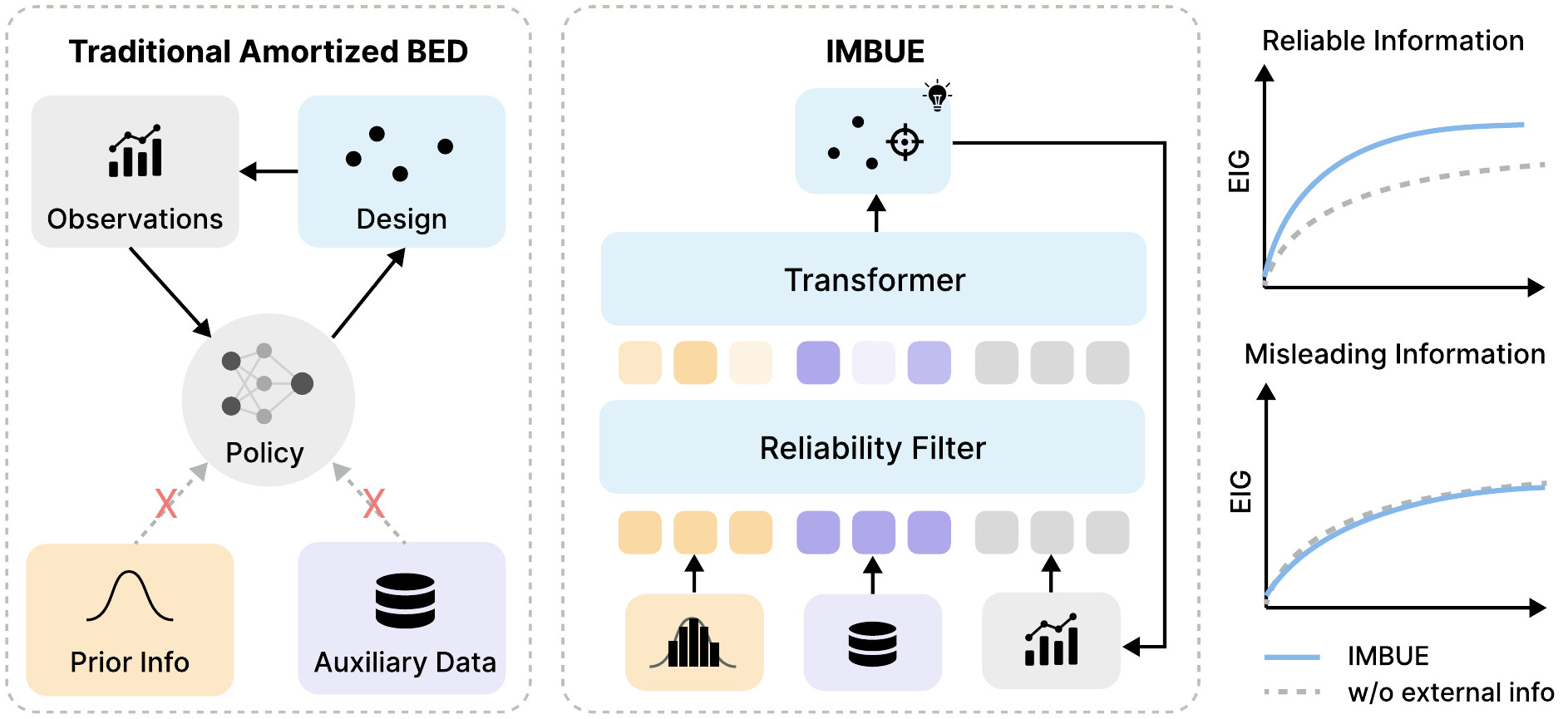}
  \caption{\textbf{Overview of \method}.
  \emph{Left}: Existing amortized BED policies fail to encode additional information available at deployment. \emph{Middle}: \method enables in-context adaptation at deployment by encoding external knowledge as additional input tokens, together with a reliability filter that excludes unreliable information. \emph{Right}: \method accelerates early information acquisition when external information is reliable, and remains robust to misleading information. }
  \vspace{-2mm}
  \label{fig:intro}
\end{figure}

A central limitation of existing amortized BED methods is that the policy $\pi_\psi$ is trained to be optimal in expectation under the training prior $p(\theta)$ with no input pathway for additional information about the specific instance being run. In practice, however, experimenters rarely approach a new study without background information \citep{mikkola2024prior}. Two forms of such information are particularly common. The first is \emph{user-specified prior knowledge} about the parameters of interest, drawn from domain expertise, published literature, or pilot studies, as when a clinician specifies an informed prior over pharmacokinetic parameters before a new trial \citep{mikkola2024prior}. The second is \emph{auxiliary observations from previous instances} of the same experiment, as in psychophysics, where data collected from earlier participants serve as informative context for designing the experiment for a new subject. Both can sharpen the effective prior for the current instance and accelerate information acquisition, but current amortized BED frameworks have no mechanism to incorporate them at~deployment.

Yet simply admitting such inputs is not enough. External information can be noisy or only partially relevant, and may come from conditions that differ from the current instance \citep{tighineanu2024scalable, li2025robust}. A user-specified prior may place probability mass in the wrong region of parameter space, and observations from previous participants may come from a population that does not match the current subject. Conditioning on such inputs without scrutiny can pull the policy away from what the accumulated observations actually support, leaving it worse off than a no-external baseline, a phenomenon known as negative transfer \citep{Rosenstein2005ToTO,li2025robust}. The challenge, then, is to design an amortized BED framework that admits external inputs at deployment without retraining while remaining robust when those inputs are unreliable.

To address these shortcomings, we introduce \textbf{I}n-context a\textbf{M}ortized \textbf{B}ED with \textbf{U}nified knowledge \textbf{E}ncoding (\method), an amortized BED framework that admits both forms of external knowledge at deployment within a shared Transformer architecture (\Cref{fig:intro}). External inputs, whether a user-specified prior or observations from previous instances of the experiment, are encoded as additional input tokens and processed in-context with the accumulated observations by the shared Transformer. A reliability filter scores each in-context token based on the current observations and excludes the tokens judged inconsistent from subsequent computation. During training, we expose the policy to in-context inputs from both reliable and unreliable regimes, so that a single trained policy can absorb whatever information is supplied at deployment. In the limit where all in-context tokens are excluded, the forward pass reduces to a history-only amortized BED policy.

We summarize our \textbf{contributions} below.
\begin{itemize}
    \item We identify a structural failure mode of runtime conditioning in amortized BED. Existing methods can be extended to accept external inputs through naive concatenation, but this baseline can degrade performance even when the inputs are informative, since the policy has no learned mechanism to determine when and how to rely on them.

    \item We propose \method, the first amortized BED framework that simultaneously supports (i) in-context knowledge available at deployment, including user-specified priors and observations from previous instances of the experiment, through a unified token interface trained under a mixture of reliable and unreliable inputs, and (ii) a learned reliability filter that assesses the relevance of each in-context token to the current experiment based on the accumulated observations, and filters out irrelevant tokens before they inform the design policy. The interface and training mixture provide the gains and most of the robustness; the filter further reduces negative transfer under conflicting inputs. Both components are realized within a single Transformer trained end-to-end.
    
    \item We evaluate \method on four standard BED benchmarks. When external knowledge is reliable, \method accelerates early-stage information acquisition. When it is not, \method reduces the degradation that naive concatenation incurs and closely matches the no-external baseline. A staged ablation separates these contributions, and \method does not degrade sharply under three deployment shifts absent from training.
\end{itemize}

\section{Related work}
\label{sec:related_work}

\textbf{Bayesian Experimental Design (BED).}
Sequential BED aims to maximize the EIG over a sequence of experiments. Traditional adaptive BED requires nested Monte Carlo or variational EIG estimation \citep{rainforth2018nesting, foster2019variational, foster2020unified} together with Bayesian posterior inference at every step \citep{chaloner1995bayesian, ryan2016review, rainforth2024modern}, which is computationally demanding. Amortized policy-based BED, introduced by Deep Adaptive Design \citep{foster2021deep}, avoids this cost by training a design policy $\pi_\psi$ offline and producing each subsequent design through a single forward pass at deployment. The framework has since been extended to implicit-likelihood models \citep{ivanova2021implicit} and trained with reinforcement learning to remove the need for differentiable simulators \citep{blau2022optimizing, blau2023cross}. More recent work couples the policy with a jointly trained inference network in fully end-to-end architectures, including ALINE \citep{huang2025aline} and JADAI \citep{bracher2025jadai}. A parallel direction, Step-DAD \citep{hedman2025stepdad}, relaxes full amortization through policy refinement at deployment that requires gradient updates and simulator access during the live experiment, which limits its applicability when designs must be produced in real time. Across all of these methods, the prior under which $\pi_\psi$ is trained is treated as a fixed property of the model, and there is no input pathway through which an experimenter can supply in-context knowledge about the current experiment at deployment. Our work addresses this gap by introducing a single design policy that accepts in-context knowledge at deployment.

\textbf{In-context knowledge in Bayesian methods.}
A separate line of work studies how in-context knowledge can shape the behavior of a Bayesian model. In Bayesian optimization (BO), this takes the form of user-specified priors over the optimum or data from related tasks \citep{souza2021bayesian, hvarfner2022pi, hvarfner2023general, huang2022bayesian, liu2024large, swersky2013multi, feurer2018scalable, wistuba2018scalable, perrone2018scalable}. Closer to our setting, recent work supplies in-context knowledge to amortized meta-learners at deployment. Informed meta-learning \citep{kobalczyk2024towards} conditions a neural process on natural-language task descriptions. The Amortized Conditioning Engine (ACE) \citep{chang2024amortized} represents user-specified priors as discretized histograms and injects them as additional input tokens, with related developments using Gaussian-mixture parameterizations \citep{whittle2025distribution} and broader distribution classes \citep{mittal2025amortized}. A complementary line replaces this token-based interface with test-time guidance \citep{yang2025priorguide}. In-context transfer methods \citep{kim2022multi, ashman2024context, li2025robust} encode auxiliary datasets as additional input tokens. These approaches show that runtime conditioning is feasible within a single forward pass, but they target passive inference rather than sequential design. Our work extends this conditioning capability to sequential design and additionally addresses the reliability of the supplied in-context knowledge through a learned reliability filter.

\section{Method}
\label{sec:method}

This section is organized as follows. \Cref{sec:setup} formalizes the setup and objective, \Cref{sec:external} the token interface, \Cref{sec:filter} the reliability filter and its supervision, and \Cref{sec:backbone} the backbone and training; notation is summarized in Appendix~\ref{app:notation}.

\subsection{Problem setup}
\label{sec:setup}
Let $\theta \in \Theta \subseteq \mathbb{R}^L$ denote the latent parameters of the system under study. Given a design $x \in \mathcal{X}$, the experiment returns an outcome $y \sim p(y \mid \theta, x)$. We consider sequential experimentation over a horizon of $T$ rounds. Starting from an empty experimental history $\mathcal{D}_0 = \emptyset$, at round $t$ a design policy $\pi_\psi$ proposes $x_t \sim \pi_\psi(\cdot \mid \mathcal{D}_{t-1})$, an outcome $y_t \sim p(y \mid \theta, x_t)$ is observed, and the history is updated to $\mathcal{D}_t = \mathcal{D}_{t-1} \cup \{(x_t, y_t)\}$. The inference goal is captured by a target variable $Z$. Standard parameter-oriented BED \citep{rainforth2024modern} corresponds to $Z = \theta$, and prediction-oriented BED sets $Z = y^\star$ at test inputs $x^\star \sim p^\star(x^\star)$ with $y^\star \sim p(y^\star \mid \theta, x^\star)$ \citep{smith2023prediction}. We call one such $T$-round experiment a \emph{rollout}.

\textbf{In-context knowledge.}
We extend the standard setup with in-context knowledge available at deployment, $C = (C_\text{prior},\ C_\text{obs})$, where $C_\text{prior}$ encodes user-specified marginal beliefs over the parameters $\theta = (\theta_1, \ldots, \theta_L)$ and $C_\text{obs} = \{\mathcal{A}_k\}_{k=1}^K$ contains $K$ sets of observations from previous instances of the experiment, with $\mathcal{A}_k = \{(x_i^{(k)}, y_i^{(k)})\}_{i=1}^{T}$. We call these the \emph{prior} and \emph{previous-instance channels}; the policy now proposes $x_t \sim \pi_\psi(\cdot \mid \mathcal{D}_{t-1}, C)$. Either component may be empty, recovering the standard BED setting in which the policy conditions only on the experimental history.

\textbf{Information objective.} We measure the utility of the policy under $C$ by the conditional information gain
\begin{equation}
    \mathcal{U}(\psi;C)=I_{p_\psi}(Z;\mathcal{D}_T\mid C) =H(Z\mid C)-H_{p_\psi}(Z\mid \mathcal{D}_T,C),
    \label{eq:eig}
\end{equation}
where $H$ denotes entropy. Here $p_\psi(Z, \mathcal{D}_T \mid C)$ is the joint distribution of target and trajectory induced by running $\pi_\psi$ for $T$ rounds under $C$ (\Cref{eq:policy-joint} in Appendix~\ref{app:proof}).
During training, $C$ is sampled from the training context distribution $p_{\mathrm{train}}(C)$, and the optimized objective is $\mathbb{E}_{C\sim p_{\mathrm{train}}}\mathcal{U}(\psi;C)$.

\textbf{Limitations of existing amortized BED.}
Existing amortized BED methods train a policy $\pi_\psi$ to maximize $\mathcal{U}(\psi; C)$ under an empty channel $C = \emptyset$, with conditioning fixed at training time. Two issues arise when in-context knowledge is available at deployment. First, $\pi_\psi$ has no input pathway through which a user-specified prior or observations from a previous instance can enter, so absorbing such information requires retraining. Even with an interface in place, priors and previous observations differ in form, length, and semantics, so a naive concatenation does not yield a unified handler for both. Second, external knowledge cannot be assumed reliable. A user-specified prior may be miscalibrated and a previous instance may belong to a different population, and conditioning on it without scrutiny can degrade performance below the no-external baseline (negative transfer~\citep{Rosenstein2005ToTO, li2025robust}). \method addresses these two issues through a unified token interface (\Cref{sec:external}) and a reliability filter (\Cref{sec:filter}), built on a shared Transformer backbone (\Cref{sec:backbone}).

\subsection{In-context token interface}
\label{sec:external}

\method represents each piece of in-context knowledge as a token, so that a single Transformer accepts either $C_\text{prior}$ or $C_\text{obs}$ at deployment without retraining. We treat the two channels separately and describe the corresponding token types in turn.

\textbf{Prior tokens.}
A user supplies marginal beliefs $\rho_l(\theta_l)$, $l \in \{1, \ldots, L\}$, over individual parameters. We discretize each $\rho_l$ into a normalized histogram over $N_B$ uniform bins covering the support of $\theta_l$,
\begin{equation}
\rho_l^{(b)} = \frac{\int_{B_b} \rho_l(u)\,du}{\sum_{b'=1}^{N_B} \int_{B_{b'}} \rho_l(u)\,du},
\qquad
\boldsymbol{\rho}_l = \bigl(\rho_l^{(1)}, \ldots, \rho_l^{(N_B)}\bigr) \in \mathbb{R}^{N_B},
\label{eq:prior-hist}
\end{equation}
where $B_b$ is the $b$-th bin. This non-parametric representation accepts arbitrary marginal beliefs, including multimodal, skewed, or truncated ones, without committing to a specific parametric family. The histogram is mapped to the model dimension $D$ by an MLP $f_\text{prior}\colon \mathbb{R}^{N_B} \to \mathbb{R}^D$, and a learnable parameter-specific embedding $\boldsymbol{\zeta}_l \in \mathbb{R}^D$ ties the resulting token to its parameter,
\begin{equation}
\mathbf{e}_l^\text{prior} = f_\text{prior}(\boldsymbol{\rho}_l) + \boldsymbol{\zeta}_l \in \mathbb{R}^D.
\label{eq:prior-token}
\end{equation}
The same $\boldsymbol{\zeta}_l$ is shared with the target token for $\theta_l$ in the backbone (\Cref{sec:backbone}), aligning the prior with its target inside the encoder. Parameters without a user-specified prior are represented by uniform histograms.

\textbf{Instance tokens.}
Each instance $\mathcal{A}_k = \{(x_i^{(k)}, y_i^{(k)})\}_{i=1}^{T}$ contains $T$ design--outcome pairs. Feeding all $K \cdot T$ raw observations into the main encoder would scale self-attention quadratically in $K \cdot T$ and couple the encoder input length to the previous-instance horizon. We instead compress each $\mathcal{A}_k$ into a single instance token $\mathbf{r}_k \in \mathbb{R}^D$ through a per-instance encoder $g_\text{aux}$, a separate Transformer that operates within each instance independently and is shared across all $K$ instances,
\begin{equation}
\mathbf{r}_k = g_\text{aux}\bigl(\mathcal{A}_k\bigr) \in \mathbb{R}^D.
\label{eq:aux-token}
\end{equation}
The compression follows summary-token schemes used for passive amortized inference \citep{ashman2024context, li2025robust}. Architectural details, including the embedding of design--outcome pairs and the summary-slot pooling in $g_\text{aux}$, are provided in Appendix~\ref{app:auxiliary}. We refer to $\mathbf{e}_l^\text{prior}$ and $\mathbf{r}_k$ collectively as \emph{in-context tokens}.

\subsection{Reliability filter}
\label{sec:filter}
External knowledge is not always reliable. A user-specified prior may concentrate mass in the wrong region of parameter space, and an instance may be drawn from a population that differs from the current one. Conditioning on such tokens without scrutiny can pull the policy away from what the accumulated observations actually support, producing negative transfer. We attach a reliability filter that decides, conditional on the observations gathered so far, whether each in-context token participates in subsequent computation.

\textbf{Mechanism.}
The reliability filter $f_\omega$ produces, for each in-context token $\mathbf{e}_j$, a scalar reliability score
\begin{equation}
\alpha_j^{(t)} = \sigma\!\bigl(f_\omega(\mathbf{e}_j;\, \mathbf{E}_t)\bigr) \in [0, 1],
\label{eq:gate-logit}
\end{equation}
where $\sigma$ is the logistic sigmoid, $\mathbf{E}_t$ denotes the embeddings of the experimental history $\mathcal{D}_t$, and $f_\omega$ is a cross-attention block \citep{vaswani2017attention} that uses $\mathbf{e}_j$ as the query and $\mathbf{E}_t$ as keys and values. We threshold $\alpha_j^{(t)}$ at $\tau = 0.5$ to obtain a binary keep-or-discard decision, and tokens with $\alpha_j^{(t)} \leq \tau$ are excluded from self-attention in the backbone via a key-padding mask. When the history is empty, there is nothing to score, and all in-context tokens are excluded until the first observation arrives. We condition $f_\omega$ on $\mathbf{E}_t$ because reliability depends on consistency with gathered observations rather than on the in-context token alone. The same token may receive different filter outputs at different rounds as $\mathcal{D}_t$ grows.

\textbf{Supervision.}
Because training rollouts are simulated (\Cref{sec:setup}), the ground-truth parameter $\theta^\star$ of the current instance is available during training and gives an oracle definition of reliability. The labels use only latent variables that standard amortized BED training already draws \citep{foster2021deep, huang2025aline}, and are neither observed nor required at deployment. For a prior token associated with parameter $\theta_l$, reliability reflects whether the histogram concentrates mass near $\theta_l^\star$. We use the log-ratio of $\rho_l^{(b^\star)}$ against a uniform reference at the bin $b^\star$ that contains $\theta_l^\star$,
\begin{equation}
r_j^\text{prior} = \sigma\!\left(\beta\log \frac{\rho_l^{(b^\star)}}{1/N_B}\right).
\label{eq:label-prior}
\end{equation}
For an instance token, reliability reflects whether the previous instance shares parameters with the current one. We use a Gaussian kernel in normalized parameter space,
\begin{equation}
r_j^\text{inst} = \exp\!\Biggl(-\kappa \sum_{l=1}^{L} \biggl(\frac{\theta_l^{(k)} - \theta_l^\star}{r_l^\text{max} - r_l^\text{min}}\biggr)^{\!2}\Biggr),
\label{eq:label-aux}
\end{equation}
where $\theta^{(k)}$ denotes the parameters that generated instance $k$, and $[r_l^\text{min}, r_l^\text{max}]$ is the support of $\theta_l$ under the prior, or a fixed finite range when the support is unbounded. The task-specific ranges are reported in Appendix~\ref{app:auxiliary}.
The parameters $\beta > 0$ and $\kappa > 0$ control label sharpness. Both labels are thresholded at $1/2$ to yield binary classification targets $c_j = \mathbb{I}[r_j > 1/2]$. For prediction-oriented BED with $Z=y^\star$, the parameter-distance label of \Cref{eq:label-aux} does not directly apply, and we use a binary variant tied to the generating regime, deferred to Appendix~\ref{app:exp-gp}.

\textbf{Mixed-regime training.}
To expose the filter to both regimes, instances are drawn from a $70/30$ relevant/irrelevant mixture during training. In the \emph{relevant} regime, $\theta^{(k)}$ is a small Gaussian perturbation of $\theta^\star$ scaled per dimension by the prior support. In the \emph{irrelevant} regime, $\theta^{(k)}$ is drawn independently of $\theta^\star$. Priors are sampled with an analogous mixture of calibrated, miscalibrated, and uniform regimes, with full procedures in Appendix~\ref{app:prior-generation} and Appendix~\ref{app:auxiliary}.

\subsection{Backbone and training}
\label{sec:backbone}

\begin{figure}
  \centering
  \includegraphics[width=\textwidth]{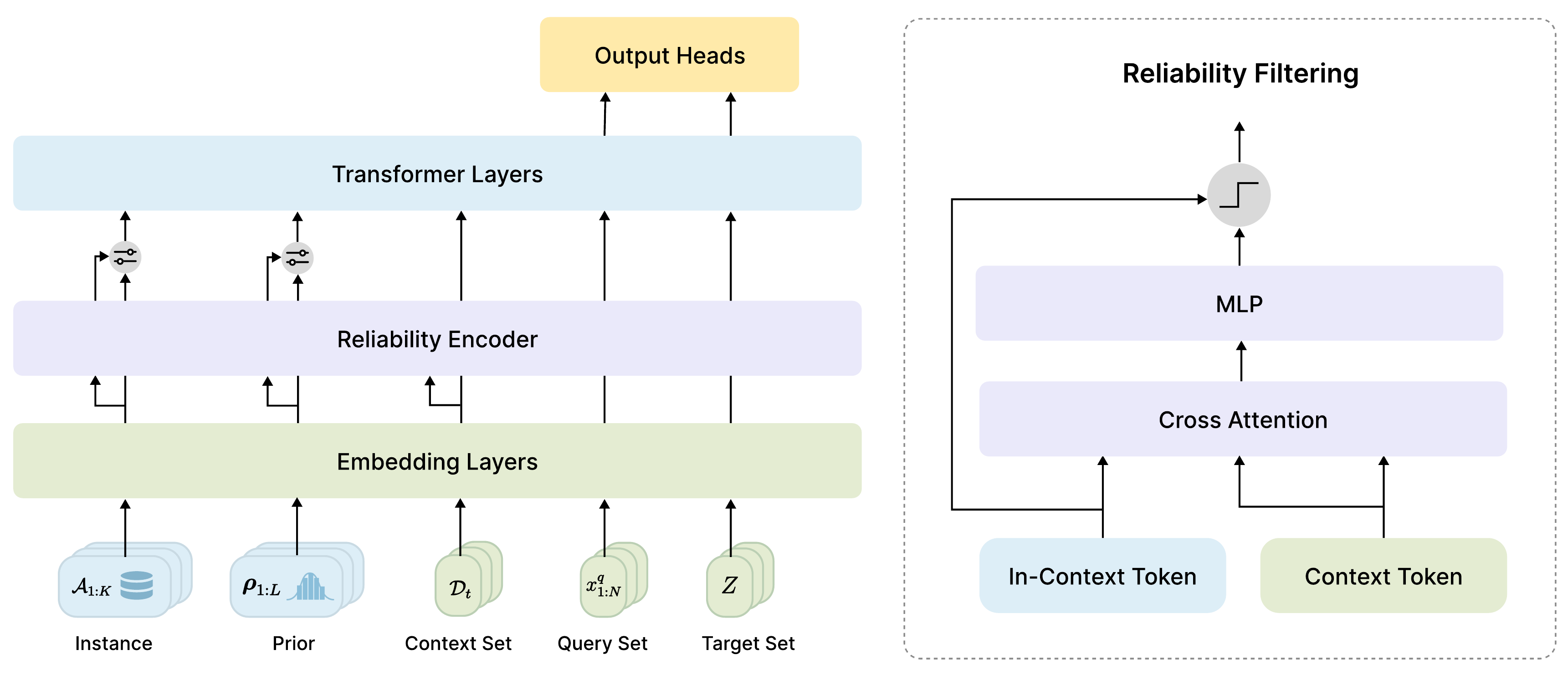}
  \caption{\textbf{Architecture of \method.} Auxiliary previous-instance data and the user-prior histogram are encoded as in-context tokens and screened by the reliability filter (right). Retained tokens are then combined with context, query, and target tokens in the shared Transformer layers.}
  \label{fig:architecture}
\end{figure}

We adopt the shared Transformer backbone, the head structure, the per-step reward, and the variational training scheme of ALINE \citep{huang2025aline} unchanged. Our contributions are the in-context token interface (\Cref{sec:external}), trained under a mixture of reliable and unreliable inputs, and the reliability filter (\Cref{sec:filter}). \Cref{fig:architecture} provides an overview of the network design; Appendix~\ref{app:architecture} provides full details.

\textbf{Variational lower bound.}
The posterior $p_\psi(Z \mid \mathcal{D}_T, C)$ entering \Cref{eq:eig} is intractable for general likelihoods, so we train against a tractable variational lower bound on $\mathcal{U}(\psi; C)$. Following \citet{barber2004algorithm}, for any variational distribution $q_\phi(Z \mid \mathcal{D}_T, C)$,
\begin{equation}
\mathcal{U}(\psi; C) \;\ge\; H(Z \mid C) + \mathbb{E}_{p_\psi(Z, \mathcal{D}_T|\, C)}\!\bigl[\log q_\phi(Z \mid \mathcal{D}_T, C)\bigr],
\label{eq:ba-bound}
\end{equation}
which is jointly maximized in $\psi$ and $\phi$ (Appendix~\ref{app:proof}). Since $H(Z \mid C)$ depends on neither $\psi$ nor $\phi$, this amounts to maximizing $\mathcal{J}(\psi, \phi; C) = \mathbb{E}_{p_\psi}[\log q_\phi(Z \mid \mathcal{D}_T, C)]$. We instantiate $q_\phi$ as an inference head on the shared Transformer encoder, trained jointly with the policy.

\textbf{Tokens and heads.}
In-context tokens that survive the filter join three further token groups inside the shared encoder. \emph{Context tokens} encode the experimental history $\mathcal{D}_t$, \emph{query tokens} encode the candidate designs at round $t$, and \emph{target tokens} are learnable slots, one per parameter when $Z = \theta$ and one per test input when $Z = y^\star$. Self-attention propagates information among the four groups under a structured attention mask, with the design head reading the query positions to produce $\pi_\psi$ and the inference head reading the target positions to produce $q_\phi$ as a Gaussian mixture over the components of $Z$ \citep{chang2024amortized}.

\textbf{Training.}
Each training iteration draws $\theta^\star$ and $C$ (\Cref{alg:context-sampling}), computes the labels $c_j$ of \Cref{sec:filter}, and simulates a $T$-round rollout in which $x_t \sim \pi_\psi(\cdot \mid \mathcal{D}_{t-1}, C)$ and $y_t \sim p(y \mid \theta^\star, x_t)$ are drawn. After round $t$, the inference head evaluates $\log q_\phi^{(t)} = \log q_\phi(Z^\star \mid \mathcal{D}_t, C)$, with $Z^\star$ the realized target ($\theta^\star$, or $y^\star$ simulated from it), and the filter re-scores the in-context tokens, giving $\alpha_j^{(t)}$ in \Cref{eq:gate-logit}. Averaged over a minibatch of rollouts, these quantities enter three losses, combined as
\begin{equation}
\mathcal{L}(\psi, \phi, \omega) = \lambda_1 \mathcal{L}_\text{policy} + \lambda_2 \mathcal{L}_\text{inf} + \lambda_3 \mathcal{L}_\text{filter},
\label{eq:total-loss}
\end{equation}
where
\begin{align}
\mathcal{L}_\text{inf} &= -\tfrac{1}{T+1} \textstyle\sum_{t=0}^{T} \log q_\phi^{(t)}, \label{eq:inf-loss}\\
\mathcal{L}_\text{policy} &= -\tfrac{1}{T}\textstyle\sum_{t=1}^{T} \gamma^{t}\, R_t \log \pi_\psi(x_t \mid \mathcal{D}_{t-1}, C), \quad \text{with } R_t = \log q_\phi^{(t)} - \log q_\phi^{(t-1)}, \label{eq:policy-loss} \\
\mathcal{L}_\text{filter} &= \tfrac{1}{T\,|C|}\textstyle\sum_{t=1}^{T}\textstyle\sum_{j=1}^{|C|}\mathrm{BCE}(\alpha_j^{(t)}, c_j).\label{eq:filter-loss}
\end{align}

\begin{itemize}[nosep, leftmargin=1.2em, labelsep=0.4em]
\item \emph{Inference loss.} $\mathcal L_{\text{inf}}$ trains the inference network $q_\phi$ to infer $Z^\star$ from the experimental history and retained in-context tokens at each round. Its final-step term optimizes \Cref{eq:ba-bound} with respect to $\phi$ and, for a fixed policy, tightens the bound. The intermediate-step terms provide accurate inference estimates for the policy rewards.
\item \emph{Policy loss.} $\mathcal L_{\text{policy}}$ trains the policy network $\pi_\psi$, which determines $\mathcal D_T$ through its selected designs. It uses the changes $R_t$ in $\log q_\phi^{(t)}$ as rewards, discounted by $\gamma \in (0, 1]$; gradients are stopped at $q_\phi$ to ensure the policy cannot raise its reward by altering the inference head. These rewards telescope to the terminal log-density minus a policy-independent initial term, thereby favoring designs that increase the variational term in \Cref{eq:ba-bound}.
\item \emph{Filter loss.} $\mathcal{L}_\text{filter}$ is the average binary cross-entropy ($\mathrm{BCE}$) of the scores $\alpha_j^{(t)}$ against the labels $c_j$ from \Cref{eq:label-prior,eq:label-aux}, where $|C|$ denotes the number of in-context tokens, and is set to $0$ when $C = \emptyset$; it is not derived from the bound and is detached from the shared embedders.

\end{itemize}

Because $R_t$ is informative only once $q_\phi$ is accurate, the first $E_\text{warm}$ epochs draw designs uniformly and optimize only $\mathcal{L}_\text{inf}$ and $\mathcal{L}_\text{filter}$; afterwards $\pi_\psi$ selects the designs and all three losses are optimized jointly. \Cref{alg:training} gives the full procedure and Appendix~\ref{app:architecture} the hyperparameters, with $\lambda_1 = \lambda_2 = \lambda_3 = 1$ and $\gamma = 1$ throughout.

\section{Experiments}
\label{sec:experiments}
We evaluate \method on four BED benchmarks. The Constant Elasticity of Substitution (CES) and Location Finding tasks support our two main claims about reliable and unreliable in-context knowledge across both channels (\Cref{sec:exp-motiv,sec:exp-location,sec:exp-ces}). \Cref{sec:exp-ablation} separates the contributions of training exposure, the token interface, and the filter. The active learning task tests whether \method filters instances individually rather than as a block (\Cref{sec:exp-gp}), and the psychometric task evaluates the previous-instance channel on a realistic scientific application (\Cref{sec:exp-psy}). Full simulator specifications, baselines, and hyperparameters are in Appendix~\ref{app:expdetails}. Baselines without a native pathway for in-context knowledge receive it through naive concatenation to the experimental history (Appendix~\ref{app:bed-baselines}), denoted by \textsuperscript{*}. Code is available at \url{https://github.com/DralnMatthew/IMBUE}.

\subsection{Naive concatenation fails}

\label{sec:exp-motiv}
\begin{wrapfigure}{R}{0.6\textwidth}
\vspace{-4mm}
\centering
\includegraphics[width=\linewidth]{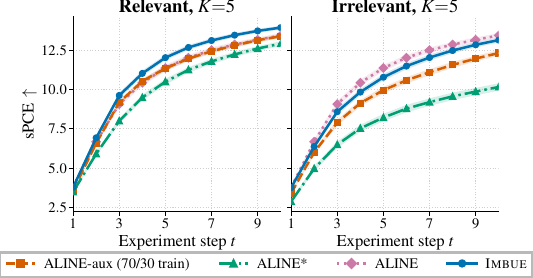}
\vspace{-4mm}
\caption{CES with $K=5$ previous instances drawn from the relevant or irrelevant regime. sPCE (mean $\pm$ 95\% confidence interval (CI)) over $T=10$ steps across $2{,}000$ rollouts.}
\label{fig:motiv}
\vspace{-4mm}
\end{wrapfigure}

The CES task \citep{arrow1961capital} infers a preference parameter $\theta\in\mathbb{R}^5$ from a respondent's comparisons of two baskets of goods over $T=10$ rounds. The target is $Z=\theta$ and the metric is the sequential Prior Contrastive Estimation (sPCE) lower bound on the total EIG \citep{foster2021deep}. We use CES in \Cref{sec:exp-motiv,sec:exp-ces} for the previous-instance channel, with $C_\text{obs}$ supplying $K$ instances from past respondents drawn from the relevant or irrelevant regime of \Cref{sec:filter}.

We first ask whether an existing amortized BED policy can make use of in-context knowledge supplied at deployment or at training time. We compare three ALINE-based variants on CES with $K=5$ instances. ALINE\textsuperscript{*} concatenates instances only at deployment, ALINE-aux is retrained on the same $70/30$ relevant/irrelevant training mixture used by \method, and ALINE is the original policy. \Cref{fig:motiv} shows that ALINE\textsuperscript{*} falls below ALINE in both regimes, indicating that previous instances unseen during training disrupt the policy regardless of their reliability. ALINE-aux roughly matches ALINE under relevant input and trails it under irrelevant input, so retraining alone extracts little from reliable instances and only partly suppresses misleading information. \method exceeds ALINE under relevant input and stays close to it under irrelevant input. The gap between ALINE-aux and \method is decomposed in \Cref{sec:exp-ablation}, and the full pattern across $K\in\{0,1,5\}$ is reported in Appendix~\ref{app:ces_results}.

\subsection{Reliable in-context knowledge accelerates acquisition}
\label{sec:exp-location}

\begin{wrapfigure}{r}{0.38\textwidth}
\vspace{-7mm}
\centering
\includegraphics[width=\linewidth]{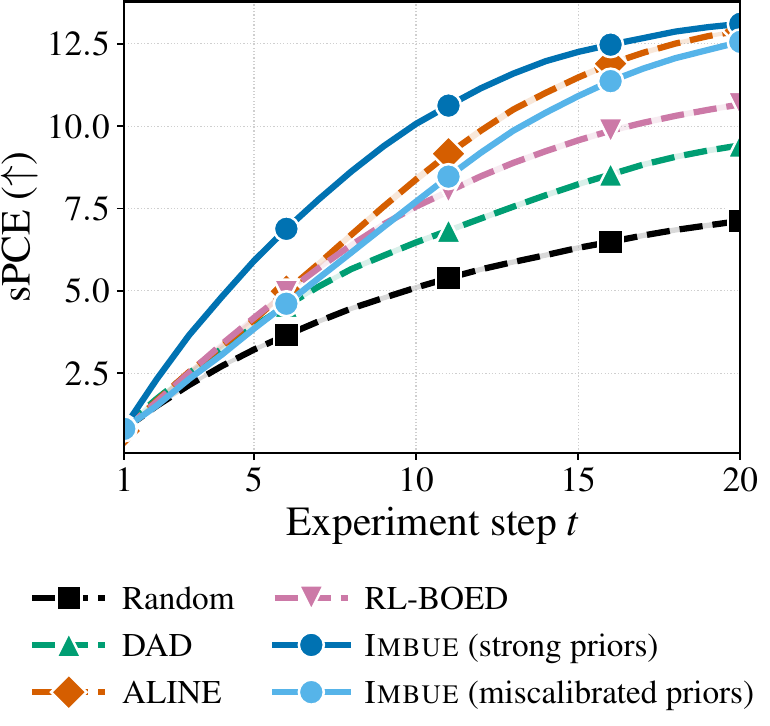}
\vspace{-3mm}
\caption{Location Finding. sPCE ($\uparrow$) over $T=20$ steps, mean $\pm$ 95\% CI across $2{,}000$ rollouts.}
\label{fig:location}
\vspace{-6mm}
\end{wrapfigure}

We then ask whether \method reaches a given level of information about $\theta$ earlier when reliable in-context knowledge is supplied, focusing on the prior channel. The Location Finding benchmark \citep{sheng2005maximum} localizes two hidden sources in $[0,1]^2$ from $T=20$ noisy measurements, with $Z=\theta$ and the sPCE metric. We supply $C_\text{prior}$ as one histogram per coordinate, and contrast a strong prior concentrated near $\theta^\star$ against a miscalibrated prior shifted away from it (Appendix~\ref{app:prior-generation}). Baselines are DAD \citep{foster2021deep}, RL-BOED \citep{blau2022optimizing}, ALINE \citep{huang2025aline}, and a random reference.

\Cref{fig:location} shows that under a strong prior, \method is consistently higher than ALINE through the early and middle rollout, and attains ALINE's final sPCE several steps earlier in the rollout. The advantage narrows toward the end as the contribution of the prior diminishes relative to the accumulated observations. Under a miscalibrated prior, \method stays close to the ALINE baseline through the rollout and finishes well above DAD, RL-BOED, and the random reference. Appendix~\ref{app:location-trajectories} reports how the per-token reliability outputs evolve as observations accumulate, along with additional results under weak and uniform priors.

The same acceleration pattern holds on the previous-instance channel. On CES, supplying \method with $K$ relevant previous instances raises its sPCE above the $K = 0$ baseline, and the gain levels off as $K$ increases (see \Cref{tab:ces-results} and the full $K$-sweep in Appendix~\ref{app:ces_results}).

\subsection{\method remains robust to unreliable in-context knowledge}
\label{sec:exp-ces}
We test whether \method reduces negative transfer on CES (\Cref{sec:exp-motiv}), supplying $K\in\{0,1,5\}$ relevant or irrelevant previous instances (\Cref{sec:filter}). Baselines are DAD\textsuperscript{*}, RL-BOED\textsuperscript{*}, ALINE\textsuperscript{*}, and a random reference.
 
\begin{table}[!ht]
\centering
\caption{CES. sPCE (mean $\pm$ 95\% CI, $2{,}000$ rollouts) under $K$ previous instances. At $K=0$ nothing is concatenated, so the $K=0$ entries of the starred methods are the standard DAD, RL-BOED, and ALINE without any in-context knowledge.}
\label{tab:ces-results}
\scalebox{0.98}{
\setlength{\tabcolsep}{5pt}
\begin{tabular}{l c c c c c}
\toprule
& $K=0$ & \multicolumn{2}{c}{$K=1$} & \multicolumn{2}{c}{$K=5$} \\
\cmidrule(lr){2-2}\cmidrule(lr){3-4}\cmidrule(lr){5-6}
Method & --- & Relevant & Irrelevant & Relevant & Irrelevant \\
\midrule
Random              & $9.05\pm 0.26$  & ---             & ---             & ---             & ---             \\
DAD\textsuperscript{*}     & $11.52\pm 0.17$ & $7.38\pm 0.21$  & $7.37\pm 0.21$  & $5.16\pm 0.21$  & $5.40\pm 0.21$  \\
RL-BOED\textsuperscript{*} & $\mathbf{14.34\pm 0.13}$ & $4.54\pm 0.20$  & $4.74\pm 0.21$  & $2.56\pm 0.15$  & $2.84\pm 0.17$  \\
% \textcolor{red}{ALINE (no in-context)} & $13.37\pm 0.15$ & $13.37\pm 0.15$ & $13.37\pm 0.15$ & $13.37\pm 0.15$ & $\mathbf{13.37\pm 0.15}$ \\
ALINE\textsuperscript{*}   & $13.37\pm 0.15$ & $13.27\pm 0.15$ & $11.10\pm 0.22$ & $13.07\pm 0.15$ & $10.16\pm 0.24$ \\
\midrule
\textbf{\method}    & $13.46\pm 0.15$ & $\mathbf{13.63\pm 0.14}$ & $\mathbf{13.42\pm 0.15}$ & $\mathbf{13.93\pm 0.12}$ & $\mathbf{13.14\pm 0.16}$ \\
\bottomrule
\end{tabular}}
\end{table}

\Cref{tab:ces-results} shows that \method's sPCE increases with $K$ when the instances are relevant, consistent with the gains from reliable prior knowledge in \Cref{sec:exp-location}. When the instances are irrelevant, \method stays close to its $K=0$ score at both values of $K$, while ALINE\textsuperscript{*} degrades as $K$ grows. DAD\textsuperscript{*} and RL-BOED\textsuperscript{*} drop under any added input regardless of its reliability. The same robustness appears on the prior channel under a miscalibrated prior (\Cref{fig:location}).

We further evaluate the trained models under three deployment shifts absent from training: biased but partially relevant previous instances, a different noise model in the previous instances, and priors whose support excludes $\theta^\star$ (Appendix~\ref{app:ood}). Across these settings, \method does not degrade sharply, consistent with the robustness observed above.

\subsection{Contributions of training, tokenization, and filtering}
\label{sec:exp-ablation}

We use a staged ablation on CES with $K=5$ previous instances (\Cref{tab:ces-ablation}): starting from ALINE\textsuperscript{*}, we successively add mixed-regime training (ALINE-aux), the instance-token interface (\method w/o filter), and reliability filtering (\method). \method w/o filter is retrained without the filter and keeps all instance tokens. Retrieve-then-condition uses ALINE-aux with only the two instances that best explain the observations collected so far (Appendix~\ref{app:bed-baselines}).

\Cref{tab:ces-ablation} shows that mixed-regime training and the instance-token interface improve mean sPCE over ALINE\textsuperscript{*} under both reliable and unreliable inputs. Adding the filter further increases mean sPCE under biased and irrelevant inputs, while performance under relevant inputs remains similar. This is consistent with the filter's purpose: limiting negative transfer from unreliable instances and simply retaining reliable ones, for which no additional gain is expected. Under irrelevant inputs, the benefit of filtering is largest in the middle of CES rollouts, grows with $K$, and is more persistent on the psychometric task (Appendix~\ref{app:ablation-and-sensitivity}). Compared with ALINE-aux, Retrieve-then-condition improves sPCE under irrelevant inputs but reduces it under relevant inputs, since its fixed top-two selection must discard useful instances. \method has no such budget and can keep all five instances or drop all five at any round, depending on the observations accumulated so far. The parameter and latency overhead of the token interface and reliability filter is reported in Appendix~\ref{app:overhead}.

\begin{table}[!ht]
\centering
\caption{Staged ablation on CES with $K=5$ previous instances. sPCE (mean $\pm$ 95\% CI, $2{,}000$ rollouts). Biased instances are a deployment shift absent from training (Appendix~\ref{app:ood}).}
\label{tab:ces-ablation}
\setlength{\tabcolsep}{5pt}
\begin{tabular}{l c c c c}
\toprule
& $K=0$ & \multicolumn{3}{c}{$K=5$} \\
\cmidrule(lr){2-2}\cmidrule(lr){3-5}
Method  & --- & Relevant & Biased & Irrelevant \\
\midrule
ALINE\textsuperscript{*}  & $13.37\pm 0.15$ & $13.07\pm 0.15$ & $11.26\pm 0.21$ & $10.16\pm 0.24$ \\
ALINE-aux & $13.19\pm 0.15$ & $13.52\pm 0.13$ & $12.72\pm 0.16$ & $12.23\pm 0.18$ \\
\method w/o filter  & $13.39\pm 0.15$ & $\mathbf{13.99\pm 0.12}$ & $13.10\pm 0.15$ & $12.97\pm 0.16$ \\
\textbf{\method} & $\mathbf{13.46\pm 0.15}$ & $13.93\pm 0.12$ & $\mathbf{13.23\pm 0.15}$ & $\mathbf{13.14\pm 0.16}$ \\
\midrule
Retrieve-then-condition  & --- & $13.29\pm 0.14$ & --- & $12.69\pm 0.17$ \\
\bottomrule
\end{tabular}
\end{table}

On the prior channel in Location Finding, however, the filter has little effect on sPCE across all four tested prior conditions (Appendix~\ref{app:ablation-and-sensitivity}), and \method w/o filter retains the robustness to miscalibrated priors observed in \Cref{fig:location}. Together with the CES and psychometric filter ablations, this indicates that filtering provides additional protection primarily against unreliable previous-instance input.

\subsection{\method handles mixed-relevance inputs per instance}
\label{sec:exp-gp}
In \Cref{sec:exp-motiv,sec:exp-ces}, $C_\text{obs}$ is supplied as either completely relevant or irrelevant. We test whether \method can retain relevant instance tokens and exclude irrelevant ones within a single forward pass. The benchmark is 1D active learning. At each step the policy chooses a design $x_t \in [-5, 5]$ and observes a noisy evaluation of a function $f^\star$. The target is $Z = y^\star$ at held-out test inputs, and the metric is the root mean squared error (RMSE) of the predictions on that test set. \emph{Synthetic} is a fresh draw from the same GP prior used during training, and probes the in-distribution case. \emph{Forrester}, \emph{Higdon}, and \emph{Gramacy \& Lee} are deterministic analytic functions, and probe out-of-distribution generalization. Further benchmark details are in Appendix~\ref{app:exp-gp}. We fix $K = 5$ previous instances and sweep the number of irrelevant ones from $0$ to $5$. Relevant instances are per-instance perturbations of $f^\star$ and irrelevant instances are unrelated 1D functions. Baselines are GP\textsuperscript{*} (a Gaussian process that pools the previous-instance observations with the experimental ones as if all were evaluations of $f^\star$), ALINE\textsuperscript{*}, and a multi-task GP (MT-GP) that treats each previous instance as a separate task and learns its correlation with the function $f^\star$ (Appendix~\ref{app:exp-gp}).

\begin{figure}[!ht]
\centering
\includegraphics[width=1.0\textwidth]{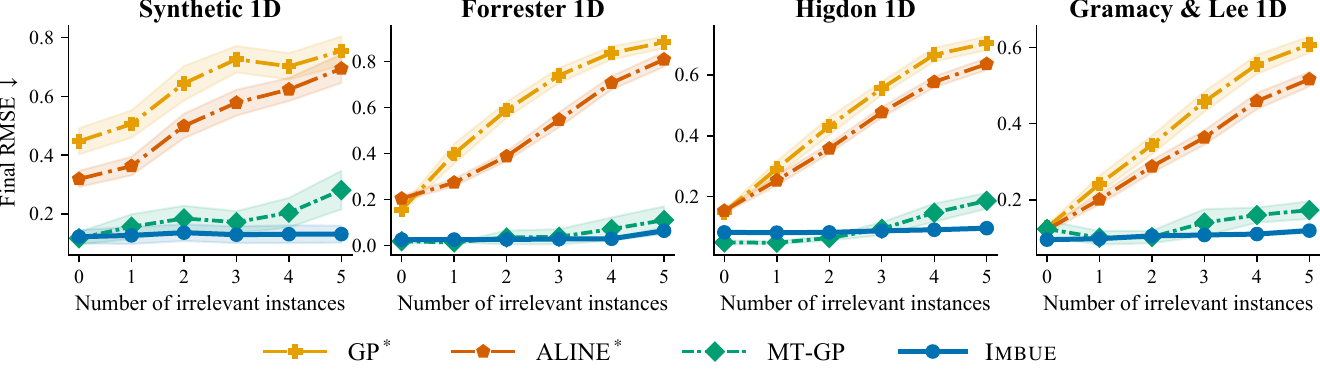}
\caption{Mixed previous-instance input on four 1D GP active learning benchmarks. Final RMSE ($\downarrow$, mean $\pm$ 95\% CI over $100$ rollouts). With $K=5$ instances, the $x$-axis varies the number that are irrelevant from $0$ to $5$.}

\label{fig:gp_comparison}
\end{figure}

\Cref{fig:gp_comparison} shows that GP\textsuperscript{*}, ALINE\textsuperscript{*}, and MT-GP degrade as the number of irrelevant instances grows, whereas \method remains essentially flat on all four benchmarks. The contrast with ALINE\textsuperscript{*}, which shares the backbone with \method but concatenates the instances at deployment, shows that the backbone architecture alone does not provide this robustness.

MT-GP and \method handle irrelevant input differently. MT-GP fuses all $K$ instances into a joint model with learned task-correlation weights. \method instead applies a binary keep-or-discard decision per token, excluding tokens judged unreliable from self-attention. RMSE curves over experimental steps, an extended baseline comparison, and an inference-time comparison are in Appendix~\ref{app:gp_results}.

\subsection{Realistic psychometric task}
\label{sec:exp-psy}
Our final experiment infers the parameters of the psychometric model \citep{watson2017quest+} from binary yes/no responses to stimulus intensities $x \in [-5, 5]$ over $T = 30$ steps. The target is $Z = \theta = (\theta_\text{threshold}, \theta_\text{slope}, \theta_\text{guess}, \theta_\text{lapse}) \in \mathbb{R}^4$. The threshold $\theta_\text{threshold}$ and slope $\theta_\text{slope}$ are the parameters of scientific interest, while the guess rate and lapse rate are nuisance parameters constrained to small physiological supports \citep{prins2013psi}. The metric is the joint RMSE on the threshold--slope marginals of the inferred posterior. We supply $C_\text{obs}$ as $K = 5$ instances from previous participants drawn from the relevant or irrelevant regime of \Cref{sec:filter}. As baselines we include the amortized policy ALINE \citep{huang2025aline} and two classical psychometric methods, QUEST+ \citep{watson2017quest+} and Psi-marginal \citep{prins2013psi}. All serve as no-external references.

\begin{wrapfigure}{r}{0.45\textwidth}
\vspace{-6mm}
\centering
\includegraphics[width=\linewidth]{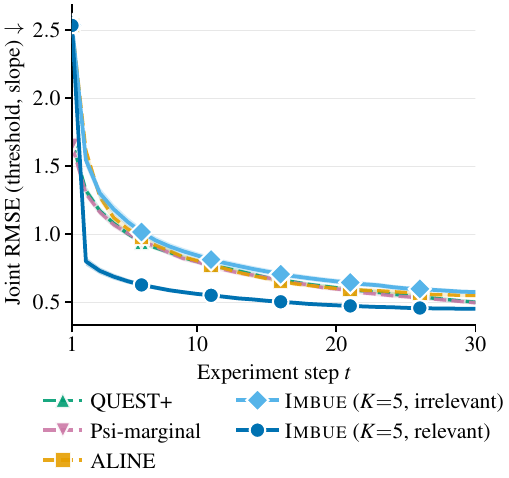}
\vspace{-3mm}
\caption{Psychometric task. Joint RMSE on $(\theta_\text{threshold},\theta_\text{slope})$ ($\downarrow$) over $T=30$ steps, mean $\pm$ 95\% CI across $500$ rollouts.}
\label{fig:psy}
% \vspace{-6mm}
\end{wrapfigure}

\Cref{fig:psy} shows that the patterns observed on CES in \Cref{sec:exp-ces} carry over to the psychometric task. \method exploits the previous-instance channel that the classical methods do not natively accept. With relevant previous instances, it attains lower joint RMSE than the no-external baselines from the second step onward; its RMSE drops sharply in the first few steps, and the gap narrows toward the end. With irrelevant previous instances, \method tracks the no-external baselines closely. Its joint RMSE is also lower than that of \method w/o filter, with an advantage that grows and persists longer as $K$ increases (Appendix~\ref{app:ablation-and-sensitivity}). Appendix~\ref{app:psychometric_results} extends this comparison to other values of $K$. \method is comparable to ALINE, QUEST+, and Psi-marginal without previous-instance input ($K=0$), and its gains with relevant instances are larger at higher $K$. ALINE\textsuperscript{*}, by contrast, does not improve over its $K=0$ RMSE even with relevant instances, and its RMSE grows with $K$ under irrelevant ones.

\section{Conclusion}
\label{sec:conclusion}
We introduced \method, an amortized BED framework that lets a trained design policy incorporate in-context knowledge supplied at deployment without retraining, either as a user-specified prior or as observations from previous instances of the experiment. A unified token interface encodes priors and observations from previous instances as additional input tokens, and a reliability filter scores each in-context token against the accumulated observations to provide an interpretable per-token decision. The staged ablation attributes the gains and most of the robustness to this interface and its mixed-regime training, with the filter adding protection when tokens conflict with the observations, mainly on the previous-instance channel. Across four standard BED benchmarks, \method accelerates early-stage information acquisition with reliable external inputs, remains close to the no-external baseline with unreliable ones, and does not degrade sharply under three deployment shifts absent from training.

\paragraph{Limitations and future work.}
The in-context knowledge we consider is structured, with priors represented as histograms and previous instances as sequences of design--outcome pairs. Priors enter as parameter-wise marginals, so parameter dependencies are not represented, and the likelihood, design, and observation spaces are assumed to stay within the trained task family (Appendix~\ref{app:ood} probes three specific deployment shifts). Admitting richer forms, such as natural-language task descriptions, would bring amortized BED closer to the settings in which domain experts naturally encode their knowledge. The main qualitative conclusions hold across the tested candidate-pool sizes and constructions (Appendix~\ref{app:pool-sensitivity}), but candidate enumeration does not scale to very high-dimensional designs; extending \method to continuous design spaces through actor-critic formulations \citep{shen2025variational} would broaden its applicability. Finally, the filter is less reliable in the first few rounds, when the limited experimental history makes it difficult to distinguish informative from misleading external inputs.

\clearpage

\begin{ack}
This work was supported by EU funding ERC ODD-ML 101201120 and the Research Council of Finland (Flagship programme: Finnish Center for Artificial Intelligence FCAI, decision 359207). The authors acknowledge the research environment provided by ELLIS Institute Finland. SK was also supported by the UKRI Turing AI World-Leading Researcher Fellowship, [EP/W002973/1]. The authors wish to thank the Aalto Science-IT project and CSC--IT Center for Science, Finland, for the computational and data storage resources provided. Views and opinions expressed are however those of the author(s) only and do not necessarily reflect those of the European Union or the European Research Council Executive Agency. Neither the European Union nor the granting authority can be held responsible for them.
\begin{center}
\includegraphics[width=0.3\linewidth]{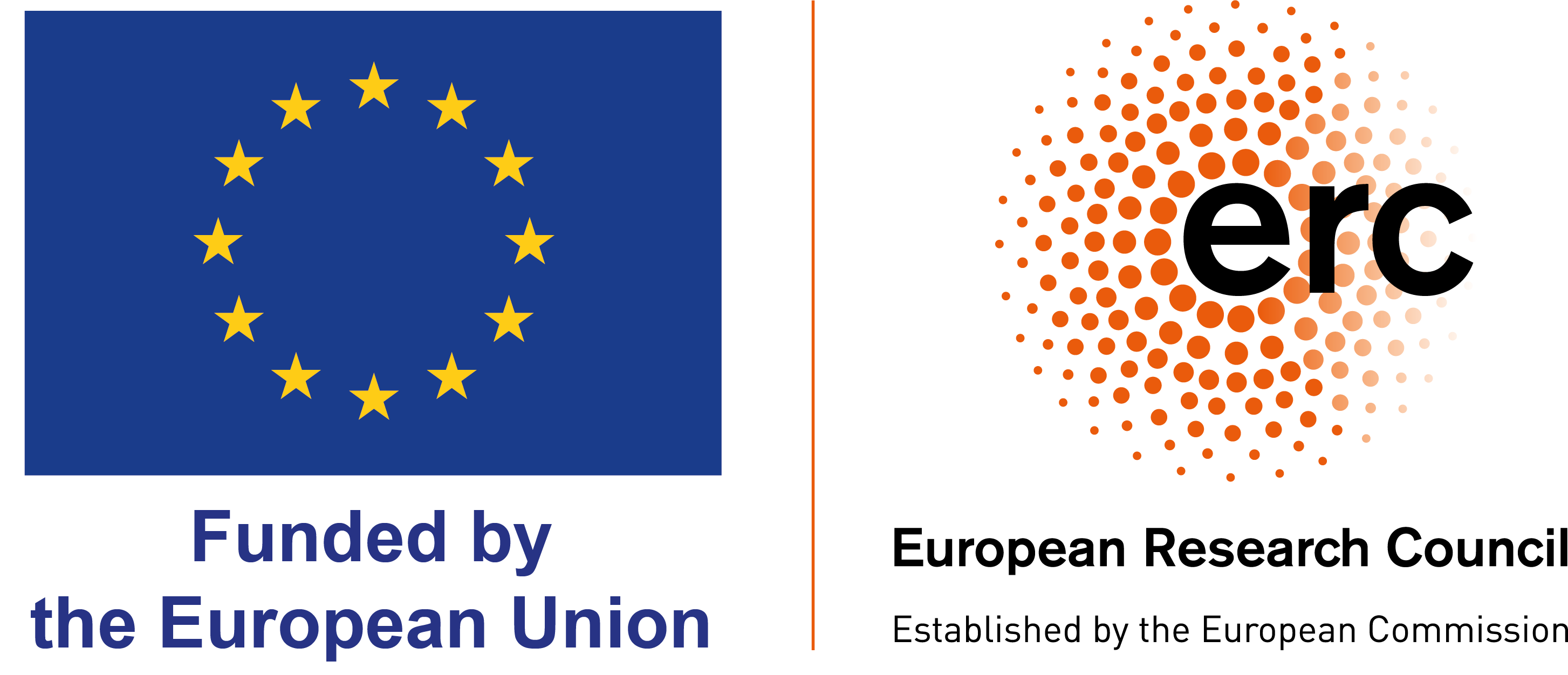}
\end{center}
\end{ack}

\bibliographystyle{unsrtnat}
\bibliography{ref}

\clearpage

\appendix

\begin{center}
    {\Large \bf Appendix}
\end{center}
\vspace{10pt}

The appendix is organized as follows.
\begin{itemize}
    \item Appendix~\ref{app:notation} summarizes the notation used throughout the paper.
    \item Appendix~\ref{app:proof} derives the Barber--Agakov bound stated in \Cref{eq:ba-bound}.
    \item Appendix~\ref{app:furdetails} provides architecture and training details, including the construction of prior and instance tokens, the supervision of the reliability filter, and the training algorithm.
    \item Appendix~\ref{app:expdetails} describes the four experimental setups, including simulator specifications, baseline implementations, and hyperparameters.
    \item Appendix~\ref{app:addexpdetails} reports additional results for different tasks, ablation and sensitivity studies on the reliability filter, deployment shifts absent from training, the sensitivity to the candidate design pool, and the computational overhead of in-context conditioning.
    \item Appendix~\ref{app:computation_resources} reports the computational resources and software dependencies used in this work.
\end{itemize}

\section{Notation}
\label{app:notation}
\Cref{tab:notation} summarizes the notation used in the main text.

\begin{table}[!ht]
\centering
\small
\caption{Notation used in the main text.}
\label{tab:notation}
\setlength{\tabcolsep}{6pt}
\renewcommand{\arraystretch}{1.05}
\begin{tabular}{l p{0.66\textwidth}}
\toprule
Symbol & Meaning \\
\midrule
\multicolumn{2}{l}{\emph{Problem setup (\Cref{sec:setup})}} \\
$\theta \in \Theta \subseteq \mathbb{R}^L$ & Latent parameters of the system; $L$ is the parameter dimension \\
$\theta^\star$ & Ground-truth parameters of a training rollout, drawn from the prior; unknown at deployment \\
$x \in \mathcal{X}$, $y$ & Design and outcome; $p(y \mid \theta, x)$ is the likelihood \\
$T$, $\mathcal{D}_t$ & Horizon of the experiment, and history $\{(x_i, y_i)\}_{i=1}^{t}$ after $t$ rounds with $\mathcal{D}_0 = \emptyset$ \\
$Z$ & Inference target: $Z = \theta$ (parameter-oriented) or $Z = y^\star$ at test inputs $x^\star \sim p^\star$ (prediction-oriented) \\
Rollout & One $T$-round sequential experiment, simulated during training \\
$\pi_\psi(x_t \mid \mathcal{D}_{t-1}, C)$ & Design policy with parameters $\psi$ \\
$q_\phi(Z \mid \mathcal{D}_t, C)$ & Variational posterior (inference network) with parameters $\phi$ \\
$p_\psi(Z, \mathcal{D}_T \mid C)$ & Joint distribution of target and trajectory induced by running $\pi_\psi$ under $C$ (\Cref{eq:policy-joint}) \\
$\mathcal{U}(\psi; C)$, $\mathcal{J}(\psi, \phi; C)$ & Information objective (\Cref{eq:eig}) and the variational term of its lower bound (\Cref{eq:ba-bound}) \\
\midrule
\multicolumn{2}{l}{\emph{In-context knowledge (\Cref{sec:external})}} \\
$C = (C_\text{prior}, C_\text{obs})$, $|C|$ & In-context knowledge, consisting of the prior channel and the previous-instance channel (either may be empty), and its number of in-context tokens ($L$ prior tokens or $K$ instance tokens) \\
$\rho_l$, $\boldsymbol{\rho}_l \in \mathbb{R}^{N_B}$ & User-specified marginal belief over $\theta_l$, and its normalized histogram over $N_B$ bins (\Cref{eq:prior-hist}) \\
$f_\text{prior}$, $\boldsymbol{\zeta}_l$, $\mathbf{e}_l^\text{prior}$ & Histogram MLP, parameter-specific embedding, and prior token (\Cref{eq:prior-token}) \\
$\mathcal{A}_k$, $K$ & Observations $\{(x_i^{(k)}, y_i^{(k)})\}_{i=1}^{T}$ of previous instance $k$, and the number of instances \\
$\theta^{(k)}$ & Parameters that generated instance $k$ \\
$g_\text{aux}$, $n_s$, $\mathbf{r}_k$ & Per-instance encoder, its number of summary slots, and the instance token (\Cref{eq:aux-token}) \\
$\mathbf{e}_j$, $D$ & An in-context token (prior or instance), and the model dimension \\
\midrule
\multicolumn{2}{l}{\emph{Reliability filter (\Cref{sec:filter})}} \\
$\mathbf{E}_t$ & Embeddings of the experimental history $\mathcal{D}_t$ \\
$f_\omega$, $\alpha_j^{(t)}$, $\tau$ & Filter with parameters $\omega$, reliability score of token $j$ (\Cref{eq:gate-logit}), and keep-or-discard threshold ($0.5$) \\
$r_j^\text{prior}$, $r_j^\text{inst}$, $c_j$ & Oracle reliability labels (\Cref{eq:label-prior,eq:label-aux}) and their binary version $c_j = \mathbb{I}[r_j > 1/2]$ \\
$b^\star$, $\beta$, $\kappa$ & Histogram bin containing $\theta_l^\star$, and the prior and instance label sharpness\\
$[r_l^\text{min}, r_l^\text{max}]$ & Support (or finite simulation range) of $\theta_l$ used for normalization \\
$p_\text{rel}$, $\Sigma_\text{rel}$ & Probability of the relevant regime ($0.7$), and covariance of the relevant-regime perturbation (standard deviation $0.1$ of the support per dimension) \\
$p_\text{good}$, $p_\text{uniform}$ & Mixture probabilities of the prior regimes during training ($p_\text{good} = 0.7$, $p_\text{uniform} = 0.2$; \Cref{alg:context-sampling}) \\
\midrule
\multicolumn{2}{l}{\emph{Training (\Cref{sec:backbone})}} \\
$\mathcal{L}_\text{policy}$, $\mathcal{L}_\text{inf}$, $\mathcal{L}_\text{filter}$ & Policy, inference, and filter losses, with weights $\lambda_1, \lambda_2, \lambda_3$ (\Cref{eq:total-loss}) \\
$R_t$, $\gamma$ & Per-step reward, and discount factor \\
$N_q$ & Size of the candidate design pool \\
$E$, $E_\text{warm}$ & Total and warm-up training epochs \\
\bottomrule
\end{tabular}
\end{table}

\section{Derivation of the Barber--Agakov bound}
\label{app:proof}

\paragraph{Proof of \Cref{eq:ba-bound}.}
Fix the in-context knowledge $C$. Under the design policy $\pi_\psi$, let $p_\psi(Z, \mathcal{D}_T \mid C)$ denote the induced joint distribution over the target and the experimental trajectory, that is, the distribution of $(Z, \mathcal{D}_T)$ obtained by running $\pi_\psi$ for $T$ rounds under $C$. For $Z = \theta$,
\begin{equation}
p_\psi(\theta, \mathcal{D}_T \mid C) = p(\theta \mid C) \prod_{t=1}^{T} \pi_\psi(x_t \mid \mathcal{D}_{t-1}, C)\, p(y_t \mid \theta, x_t),
\label{eq:policy-joint}
\end{equation}
where $p(\theta \mid C)$ is the distribution of the parameters given $C$ (\Cref{alg:context-sampling} specifies it during training); for $Z = y^\star$ the analogous form additionally draws $x^\star \sim p^\star(x^\star)$ and $y^\star \sim p(y^\star \mid \theta, x^\star)$. The conditional information gain equals
\[
    \mathcal{U}(\psi; C)
    = H(Z \mid C) - H_{p_\psi}(Z \mid \mathcal{D}_T, C),
\]
where $H(Z \mid C)$ depends only on the prior $p(Z \mid C)$ and $H_{p_\psi}(Z \mid \mathcal{D}_T, C)$ is taken under the policy-induced joint $p_\psi(Z, \mathcal{D}_T \mid C)$. Writing the conditional entropy as an expectation,
\[
    H_{p_\psi}(Z \mid \mathcal{D}_T, C)
    = -\mathbb{E}_{p_\psi(Z, \mathcal{D}_T \mid C)}\!\bigl[
        \log p_\psi(Z \mid \mathcal{D}_T, C)
    \bigr],
\]
we obtain
\[
    \mathcal{U}(\psi; C)
    = H(Z \mid C)
    + \mathbb{E}_{p_\psi(Z, \mathcal{D}_T \mid C)}\!\bigl[
        \log p_\psi(Z \mid \mathcal{D}_T, C)
    \bigr].
\]
For any variational distribution $q_\phi(Z \mid \mathcal{D}_T, C)$, the expected Kullback--Leibler divergence is non-negative,
\[
\begin{aligned}
    0
    &\leq \mathbb{E}_{p_\psi(\mathcal{D}_T \mid C)}\!\Bigl[
        \mathrm{KL}\!\bigl(
            p_\psi(Z \mid \mathcal{D}_T, C) \,\|\, q_\phi(Z \mid \mathcal{D}_T, C)
        \bigr)
    \Bigr] \\
    &= \mathbb{E}_{p_\psi(Z, \mathcal{D}_T \mid C)}\!\bigl[
        \log p_\psi(Z \mid \mathcal{D}_T, C) - \log q_\phi(Z \mid \mathcal{D}_T, C)
    \bigr],
\end{aligned}
\]
so that
\[
    \mathbb{E}_{p_\psi(Z, \mathcal{D}_T \mid C)}\!\bigl[
        \log p_\psi(Z \mid \mathcal{D}_T, C)
    \bigr]
    \geq \mathbb{E}_{p_\psi(Z, \mathcal{D}_T \mid C)}\!\bigl[
        \log q_\phi(Z \mid \mathcal{D}_T, C)
    \bigr].
\]
Substituting this inequality into the mutual-information identity yields
\[
    \mathcal{U}(\psi; C)
    \geq H(Z \mid C)
    + \mathbb{E}_{p_\psi(Z, \mathcal{D}_T \mid C)}\!\bigl[
        \log q_\phi(Z \mid \mathcal{D}_T, C)
    \bigr],
\]
which is the bound in \Cref{eq:ba-bound}.

\section{Further details}
\label{app:furdetails}

\subsection{Training algorithm}
\Cref{alg:training} summarizes the joint training procedure for the design policy $\pi_\psi$, the inference network $q_\phi$, and the reliability filter $f_\omega$ described in \Cref{sec:backbone}. \Cref{alg:context-sampling} details the sampling of the in-context knowledge $C$ and the ground-truth $\theta^\star$ used in each rollout. The pseudocode presents the empty-history case. For GP active learning, initialize $\mathcal{D}_0$ with one randomly sampled target observation instead.

\begin{algorithm}[!ht]
\caption{Training procedure for \method.}
\label{alg:training}
\begin{algorithmic}[1]
\REQUIRE prior $p(\theta)$; simulator $p(y \mid \theta, x)$; candidate pool size $N_q^\text{train}$; rollout length $T$; warm-up epochs $E_\text{warm}$; total epochs $E$; loss coefficients $\lambda_1, \lambda_2, \lambda_3$
\ENSURE trained parameters $(\psi, \phi, \omega)$
\FOR{epoch $= 1, \ldots, E$}
    \FOR{each minibatch}
        \STATE Sample in-context knowledge $C$ and ground-truth $\theta^\star$ following \Cref{alg:context-sampling}
        \STATE Compute reliability labels $\{c_j\}$ for tokens in $C$ \hfill(\Cref{eq:label-prior,eq:label-aux})
        \STATE Initialize $\mathcal{D}_0 \gets \emptyset$ and $\log q_\phi^{(0)} \gets \log q_\phi(Z^\star \mid \mathcal{D}_0, C)$
        \FOR{$t = 0, 1, \ldots, T - 1$}
            \IF{epoch $\leq E_\text{warm}$}
                \STATE Sample $x_{t+1}$ uniformly from a pool of size $N_q^\text{train}$
            \ELSE
                \STATE Sample $x_{t+1} \sim \pi_\psi(\cdot \mid \mathcal{D}_t, C)$ from a pool of size $N_q^\text{train}$
            \ENDIF
            \STATE Observe $y_{t+1} \sim p(y \mid \theta^\star, x_{t+1})$
            \STATE Update $\mathcal{D}_{t+1} \gets \mathcal{D}_t \cup \{(x_{t+1}, y_{t+1})\}$
            \STATE Compute in-context token embeddings and
            reliability scores $\{\alpha_j^{(t+1)}\}$
            \hfill(\Cref{eq:gate-logit})
            \STATE Exclude in-context tokens with
            $\alpha_j^{(t+1)} \leq \tau$
            \STATE Compute $\log q_\phi^{(t+1)} \gets \log q_\phi(Z^\star \mid \mathcal{D}_{t+1}, C)$
            \STATE Compute reward $R_{t+1} \gets \log q_\phi^{(t+1)} - \log q_\phi^{(t)}$ 
        \ENDFOR
        \STATE Compute $\mathcal{L}_\text{inf}$ from $\{\log q_\phi^{(t)}\}_{t=0}^{T}$ \hfill(\Cref{eq:inf-loss})
        \STATE Compute $\mathcal{L}_\text{filter}$ from $\{\alpha_j^{(t)}\}_{t=1}^T$ and $\{c_j\}$ \hfill(\Cref{eq:filter-loss})
        \IF{epoch $\leq E_\text{warm}$}
            \STATE $\mathcal{L} \gets \lambda_2\, \mathcal{L}_\text{inf} + \lambda_3\, \mathcal{L}_\text{filter}$
        \ELSE
            \STATE Compute $\mathcal{L}_\text{policy}$ from $\{R_t\}_{t=1}^T$ and $\{\log \pi_\psi(x_t \mid \mathcal{D}_{t-1}, C)\}_{t=1}^T$ \hfill(\Cref{eq:policy-loss})
            \STATE $\mathcal{L} \gets \lambda_1\, \mathcal{L}_\text{policy} + \lambda_2\, \mathcal{L}_\text{inf} + \lambda_3\, \mathcal{L}_\text{filter}$ \hfill(\Cref{eq:total-loss})
        \ENDIF
        \STATE Update $(\psi, \phi, \omega)$ by an AdamW step on $\nabla \mathcal{L}$
    \ENDFOR
\ENDFOR
\end{algorithmic}
\end{algorithm}

\begin{algorithm}[!ht]
\caption{Sampling in-context knowledge $C$ and ground-truth $\theta^\star$ during training.}
\label{alg:context-sampling}
\begin{algorithmic}[1]
\REQUIRE prior $p(\theta)$; mode $\in \{\text{prior}, \text{previous instances}\}$; number of previous instances $K \sim \mathrm{Unif}\{0, \ldots, K_{\max}\}$
\IF{mode is \text{prior}}
    \STATE Draw the prior regime: uniform with probability $p_\text{uniform}$, Gaussian otherwise
    \STATE Sample a marginal $\rho_l$ for each parameter $l$ under the drawn regime \hfill(Appendix~\ref{app:prior-generation})
    \STATE Discretize each $\rho_l$ into a histogram $\boldsymbol{\rho}_l$ over $N_B$ bins \hfill(\Cref{eq:prior-hist})
    \FOR{each parameter $l = 1, \ldots, L$}
        \STATE Sample $\theta_l^\star \sim \boldsymbol{\rho}_l$ (calibrated) with probability $p_\text{good}$; otherwise displace the draw away from $\rho_l$ (miscalibrated) \hfill(\Cref{eq:app-prior-shift})
    \ENDFOR
    \STATE Encode each $\boldsymbol{\rho}_l$ into a prior token $\mathbf{e}_l^\text{prior}$ \hfill(\Cref{eq:prior-token})
    \STATE Set $C \gets \{\mathbf{e}_l^\text{prior}\}_{l=1}^L$

\ELSE
    \STATE Sample $\theta^\star \sim p(\theta)$
    \FOR{$k = 1, \ldots, K$}
        \STATE Draw the regime, relevant with probability $p_\text{rel}$ and irrelevant otherwise
        \IF{relevant}
            \STATE Sample $\theta^{(k)} = \theta^\star + \eta_k$ with $\eta_k \sim \mathcal{N}(0, \Sigma_\text{rel})$
        \ELSE
            \STATE Sample $\theta^{(k)} \sim p(\theta)$ independently of $\theta^\star$
        \ENDIF
        \STATE Sample designs $x_1^{(k)}, \ldots, x_T^{(k)}$ uniformly from $\mathcal{X}$
        \STATE Sample observations $y_i^{(k)} \sim p(y \mid \theta^{(k)}, x_i^{(k)})$
        \STATE Encode $\mathcal{A}_k = \{(x_i^{(k)}, y_i^{(k)})\}_{i=1}^T$ into an instance token $\mathbf{r}_k$ \hfill(\Cref{eq:aux-token})
    \ENDFOR
    \STATE Set $C \gets \{\mathbf{r}_k\}_{k=1}^K$
\ENDIF
\RETURN $C, \theta^\star$
\end{algorithmic}
\end{algorithm}

\subsection{Architecture and training details}
\label{app:architecture}

\paragraph{Token embedders.}
Designs $x$ are mapped to the model dimension $D$ by a two-layer MLP $f_x$ with feedforward dimension $128$ and output dimension $D = 32$, and observations $y$ by an analogous MLP $f_y$. Each context token combines the design and outcome embeddings, $\mathbf{e}_n^\text{ctx} = f_x(x_n) + f_y(y_n)$, and each query token uses the design embedder alone, $\mathbf{e}_m^\text{qry} = f_x(x_m^q)$. The construction of target tokens depends on the target type. For parameter targets $Z = \theta$, the $L$ target tokens $\{\boldsymbol{\zeta}_l\}_{l=1}^L$ are learnable embeddings, one per parameter dimension. These are the same $\boldsymbol{\zeta}_l$ that bind prior tokens to their parameters in \Cref{eq:prior-token}. For predictive targets $Z = y^\star$ at test inputs $\{x_m^\star\}_{m=1}^M$, each target token is the embedding of the corresponding test input, $\mathbf{e}_m^\text{tgt} = f_x(x_m^\star)$, sharing the design embedder with the context and query tokens. In-context tokens are constructed as described in Appendix~\ref{app:prior} for prior tokens and in Appendix~\ref{app:auxiliary} for instance tokens. All four token groups share the embedding dimension $D = 32$.

\begin{figure}[!ht]
\centering
\includegraphics[width=1.0\textwidth]{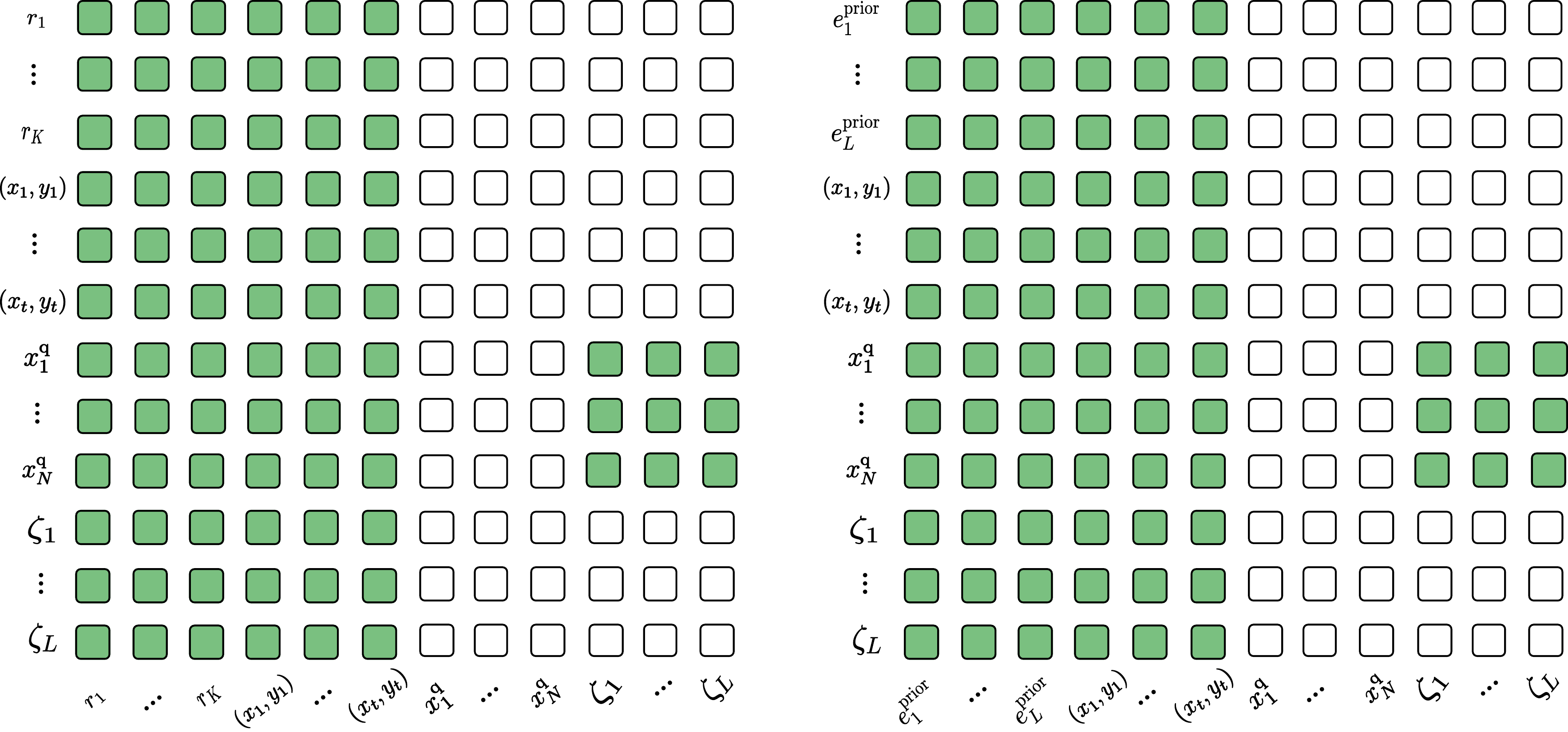}
\caption{Attention masks of \method's shared Transformer layers for the previous-instance channel (left, in-context tokens $\mathbf{r}_1,\ldots,\mathbf{r}_K$) and the prior channel (right, in-context tokens $\mathbf{e}^\text{prior}_1,\ldots,\mathbf{e}^\text{prior}_L$). Rows index queries and columns index keys, ordered as in-context tokens, context tokens $(x_i,y_i)$, query tokens $x^\text{q}_m$, and target tokens $\boldsymbol{\zeta}_l$. Shaded squares indicate allowed attention.}
\label{fig:attention}
\end{figure}

\paragraph{Shared encoder.}
The shared encoder is a $3$-layer Transformer with $4$ attention heads per layer and feedforward dimension $128$. Information flow among the four token groups is controlled by the structured attention mask described in \Cref{sec:backbone} (\Cref{fig:attention}), together with a per-example key-padding mask produced by the reliability filter (\Cref{sec:filter}) that excludes in-context tokens with $\alpha_j^{(t)} \leq \tau = 0.5$.

\paragraph{Reliability filter.}
The filter $f_\omega$ takes an in-context token as the query and the context tokens as keys and values, all detached from the main computation graph (stop-gradient) so that gradients from the filter loss do not propagate into the embedders shared with the rest of the network. Query and key-value pairs are processed by a Transformer-style cross-attention block consisting of multi-head cross-attention with $4$ heads, an Add\&LayerNorm, a position-wise feedforward network (linear, ReLU, dropout, linear), and a second Add\&LayerNorm. The resulting representation is passed through a two-layer scoring MLP with hidden dimension $16$ to produce the scalar logit, which is passed through a sigmoid to give the reliability score $\alpha_j$ in \Cref{eq:gate-logit}.

\paragraph{Output heads.}
The inference head parameterizes a Gaussian mixture with $10$ components, produced by two-layer MLPs each of hidden dimension $128$. For each component, the MLP outputs a raw weight, mean, and standard deviation. Weights are normalized through a softmax, standard deviations are passed through softplus with a floor $\sigma_\text{min} = 10^{-4}$, and means are unconstrained. The design head is a two-layer MLP with hidden dimension $128$ that maps each query position to a scalar logit, followed by a softmax over the $N_q$ candidates to form $\pi_\psi$.

\paragraph{Optimization.}

We use AdamW with initial learning rate $10^{-3}$ and the default weight decay $10^{-2}$. The learning rate follows a cosine annealing schedule that decays to zero over the full training budget. Training follows the two-stage schedule of \Cref{sec:backbone}. The first stage runs for $E_\text{warm}$ epochs with the policy gradient disabled and designs drawn uniformly from a candidate pool of size $N_q^\text{train}$. The second stage activates the full joint objective in \Cref{eq:total-loss} on the same pool. Loss coefficients are fixed at $\lambda_1 = \lambda_2 = \lambda_3 = 1$ across all benchmarks. Benchmark-specific values of $E_\text{warm}$, $N_q^\text{train}$, $T$, and other task-dependent hyperparameters are listed in \Cref{tab:hyperparams}.

\begin{table}[!ht]
\centering
\caption{Benchmark-specific hyperparameters for \method.}
\label{tab:hyperparams}
\scalebox{0.85}{
\setlength{\tabcolsep}{6pt}
\renewcommand{\arraystretch}{1.1}
\begin{tabular}{l c c c c}
\toprule
& Location Finding & CES & GP active learning & Psychometric \\
\midrule
\multicolumn{5}{l}{\emph{Experiment setup}} \\
Rollout length $T$                          & 20 & 10 & 15 & 30 \\
Parameter dimension $L$                     & 4 & 5 & --- & 4 \\
Target type $Z$                             & $\theta$ & $\theta$ & $y^\star$ & $\theta$ \\
\midrule
\multicolumn{5}{l}{\emph{Training schedule}} \\
Training-time candidate pool size $N_q^\text{train}$ & 200 & 200 & 200 & 200 \\
Total epochs $E$                            & 70,000 & 250,000 & 250,000 & 120,000 \\
Warm-up epochs $E_\text{warm}$              & 15,000 & 30,000 & 30,000 & 15,000 \\
Batch size                                  & 200 & 200 & 200 & 200 \\
Discount factor $\gamma$  & 1 & 1 & 1 & 1
\\
\midrule

\multicolumn{5}{l}{\emph{Evaluation schedule}} \\
Evaluation-time candidate pool size $N_q^\text{eval}$                           & 2,000 & 2,000 & 500 & 200 \\
Contrastive samples      & $10^6$ & $10^7$ & --- & --- \\
Evaluation-time target set size & 4 & 5 & 100 & 4 \\

\midrule
\multicolumn{5}{l}{\emph{In-context knowledge}} \\
Number of histogram bins $N_B$              & $100$ & --- & --- & --- \\
Number of summary slots $n_s$               & --- & $2$ & $4$ & $4$ \\
Maximum $K_{\max}$ at training                     & --- & 5 & 5 & 5 \\
\midrule
\multicolumn{5}{l}{\emph{Reliability filter}} \\
Threshold $\tau$                            & $0.5$ & $0.5$ & $0.5$ & $0.5$ \\
Instance label sharpness $\kappa$           & --- & 5 & --- & 4 \\
\bottomrule
\end{tabular}}
\end{table}

\subsection{Prior tokens}
\label{app:prior}
This section describes how prior tokens are generated, embedded, and labeled. The histogram representation follows ACE \citep{chang2024amortized}.

\paragraph{Histogram representation and embedding.}
For each parameter $\theta_l$, the user-supplied marginal $\rho_l$ is discretized into a histogram over $N_B = 100$ uniformly spaced bins covering the support of $\theta_l$, normalized as in \Cref{eq:prior-hist}. The histogram is mapped to the model dimension by a two-layer MLP $f_\text{prior}\colon \mathbb{R}^{N_B} \to \mathbb{R}^D$ and added to the parameter-specific embedding $\boldsymbol{\zeta}_l$ shared with the target token,
\begin{equation}
\mathbf{e}_l^\text{prior} = f_\text{prior}(\boldsymbol{\rho}_l) + \boldsymbol{\zeta}_l \in \mathbb{R}^D.
\label{eq:app-prior-token}
\end{equation}

The choice of $N_B = 100$ was not tuned in our experiments.

\paragraph{Prior generation during training.}
\label{app:prior-generation}
Each rollout draws one of two regimes.

\textit{Uniform regime} (probability $p_\text{uniform} = 0.2$). Every marginal is flat on its support, $\rho_l = \mathcal{U}(r_l^\text{min}, r_l^\text{max})$. The tokens carry no information about $\theta^\star$, and \Cref{eq:label-prior} assigns them $r^\text{prior} = 1/2$ regardless of where $\theta^\star$ falls.

\textit{Gaussian regime} (probability $1 - p_\text{uniform} = 0.8$). Each marginal is a Gaussian $\rho_l = \mathcal{N}(\mu_l, \sigma_l^2)$ with a random location $\mu_l \sim \mathcal{U}(r_l^\text{min}, r_l^\text{max})$ and a random width $\sigma_l \sim \mathcal{U}(\sigma^\text{min}, \sigma^\text{max}) \cdot (r_l^\text{max} - r_l^\text{min})$. For the Location Finding task, we use $\sigma^\text{min}=0.05$ and $\sigma^\text{max}=0.3$.

\begin{figure}[!ht]
\centering
\includegraphics[width=1.0\textwidth]{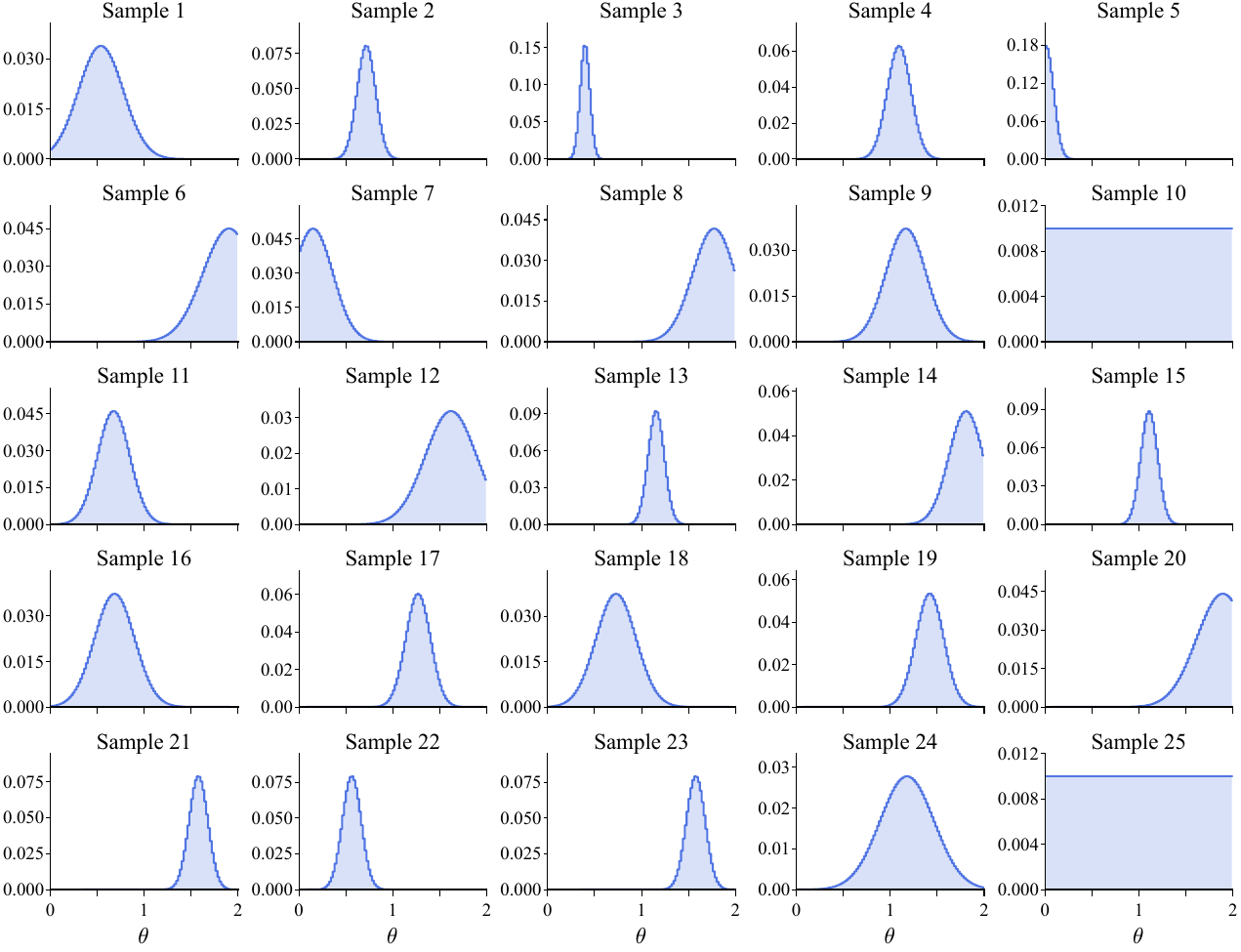}
\caption{Illustrative examples of randomly sampled prior histograms, shown on a generic support $[0, 2]$; the Location Finding experiments use the support $[0, 1]$.}
\label{fig:lf_histogram}
\end{figure}

In both regimes, $\theta^\star$ is then drawn parameter-wise from the histograms. With probability $p_\text{good}$ a parameter is \emph{calibrated}: $\theta_l^\star \sim \boldsymbol{\rho}_l$, drawn as a bin from the histogram followed by a uniform draw within that bin, so that the prior places mass near the truth. Otherwise it is \emph{miscalibrated}: the same draw $\theta_l^\text{cal}$ is displaced by a random fraction of the support toward its far side,
\begin{equation}
\theta_l^\star = \mathrm{clip}\bigl(\theta_l^\text{cal} + d_l s_l,\; r_l^\text{min},\; r_l^\text{max}\bigr),
\qquad
s_l \sim \mathcal{U}(0.3,\, 0.5) \cdot (r_l^\text{max} - r_l^\text{min}),
\label{eq:app-prior-shift}
\end{equation}
where $d_l = +1$ if $\theta_l^\text{cal}$ lies below the midpoint of $[r_l^\text{min}, r_l^\text{max}]$ and $d_l = -1$ otherwise. 
\Cref{fig:lf_histogram} shows examples of sampled priors.

\paragraph{Evaluation-time prior conditions.}
\label{app:prior-eval}
At evaluation time we probe \method's behavior under four prior conditions that span the reliability spectrum: three Gaussian priors, each with a concentration chosen from within the training range, and a uniform prior.
\begin{itemize}
\item \emph{Strong prior.} A Gaussian with $\mu_l$ centered near $\theta^\star$, with $\sigma_l = 0.1 \cdot (r_l^\text{max} - r_l^\text{min})$, modeling a confident and accurate expert. On the Location Finding support this corresponds to $\sigma = 0.1$.
\item \emph{Weak prior.} The same form as the strong prior but with a wider $\sigma_l = 0.25 \cdot (r_l^\text{max} - r_l^\text{min})$, modeling a less confident expert who still places probability mass near $\theta^\star$ but spreads it more broadly. On the Location Finding support this corresponds to $\sigma = 0.25$.
\item \emph{Miscalibrated prior.} A shifted Gaussian following the training miscalibrated sub-regime, with concentration matched to the strong prior so that the comparison isolates the effect of an incorrect mean from the level of confidence.
\item \emph{Uniform prior.} A flat histogram, modeling an expert who provides no information about $\theta^\star$ beyond its support. Under this condition the reliability filter should mask out the prior tokens, so that \method approximates the no-external baseline.
\end{itemize}
All four evaluation conditions fall within the training-time mixture described above. The main text reports results under the strong and miscalibrated regimes, and \Cref{fig:lf-prior-comparison} reports weak and uniform prior results.

\begin{figure}[!ht]
\centering
\includegraphics[width=1.0\textwidth]{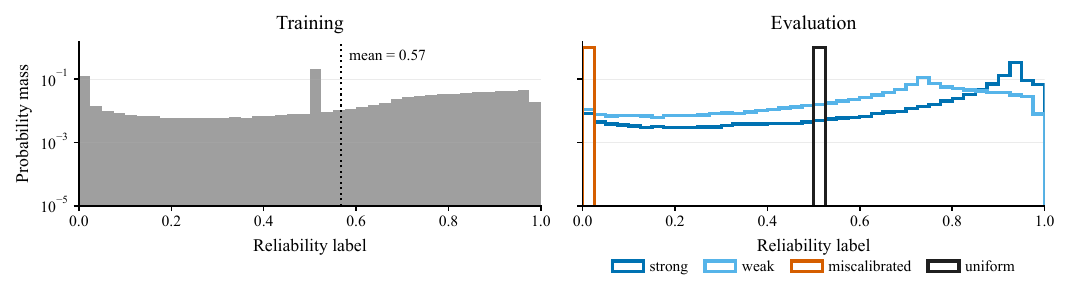}
\caption{Distribution of prior reliability labels on the Location Finding task. Training (left) versus the four evaluation regimes (right), shown as normalized densities on a log $y$-axis. The training distribution covers the full range, while each evaluation regime concentrates in the expected region.}
\label{fig:lf_reliability_hist}
\end{figure}

\paragraph{Reliability label distribution.}
\Cref{fig:lf_reliability_hist} compares the reliability labels $r_j^\text{prior}$ produced during training with those induced by the four evaluation conditions on the Location Finding task. These are the scalar labels of \Cref{eq:label-prior} with $\beta = 2$; $\beta$ only sharpens them toward $0$ and $1$ and leaves the binary targets $c_j$ unchanged. The training distribution covers both reliable and unreliable prior tokens, and the evaluation conditions concentrate in the expected regions. Strong priors concentrate near label $1$, miscalibrated priors near label $0$, weak priors at intermediate values, and uniform priors near $1/2$. The evaluation reliability is therefore in distribution for the filter.

\subsection{Instance tokens}
\label{app:auxiliary}
This section describes how instance tokens are generated and how their reliability labels are computed.
 
\paragraph{Representation.}
The previous-instance channel is a collection of $K$ instances,
\begin{equation}
C_\text{obs} = \{\mathcal{A}_k\}_{k=1}^K, \qquad \mathcal{A}_k = \{(x_i^{(k)}, y_i^{(k)})\}_{i=1}^T,
\label{eq:app-aux}
\end{equation}
each generated from the same simulator as the target experiment but under a different parameter $\theta^{(k)}$. The horizon $T$ matches the target experiment.

\paragraph{Generation during training.}
For each rollout, $\theta^\star$ is sampled from $p(\theta)$ and each of the $K$ instances $\mathcal{A}_k$ is independently assigned to one of two regimes.

\textit{Relevant regime.} The instance parameter is a perturbation of $\theta^\star$,
\begin{equation}
\theta^{(k)} = \theta^\star + \eta_k, \qquad \eta_k \sim \mathcal{N}(0, \Sigma_\text{rel}),
\label{eq:app-relevant}
\end{equation}
with $\Sigma_\text{rel} = \mathrm{diag}(\sigma_{\text{rel},l}^2)$ and $\sigma_{\text{rel},l} = 0.1\,(r_l^\text{max} - r_l^\text{min})$.

\textit{Irrelevant regime.} The instance parameter is drawn independently of $\theta^\star$, $\theta^{(k)} \sim p(\theta)$.

Conditional on $\theta^{(k)}$, designs $x_1^{(k)}, \ldots, x_T^{(k)}$ are sampled uniformly from $\mathcal{X}$ and outcomes are simulated from $p(y \mid \theta^{(k)}, x_i^{(k)})$. Random designs reflect the assumption that previous instances were collected by experimenters whose design choices need not align with ours.

\paragraph{Per-instance encoder.}
We detail the implementation of the per-instance encoder $g_\text{aux}$ defined in \Cref{eq:aux-token}. Each design--outcome pair is embedded as $\mathbf{u}_i^{(k)} = f_x(x_i^{(k)}) + f_y(y_i^{(k)}) \in \mathbb{R}^D$, using the same MLPs $f_x$ and $f_y$ as the context tokens (Appendix~\ref{app:architecture}). We prepend $n_s$ learnable summary slots $\mathbf{S} = (\mathbf{s}_1, \ldots, \mathbf{s}_{n_s})$ to these embeddings and pass the joint sequence through an $n_\text{aux}$-layer Transformer. Let $\mathbf{h}_s^{(k)} \in \mathbb{R}^D$ denote its output at summary slot $s$ for the $k$-th instance. The complete encoder mapping includes mean pooling over these slot outputs,
\begin{equation}
\mathbf{r}_k = g_\text{aux}\bigl(\mathcal{A}_k\bigr) = \frac{1}{n_s} \sum_{s=1}^{n_s} \mathbf{h}_s^{(k)} \in \mathbb{R}^D.
\label{eq:app-instance-token}
\end{equation}
We use $n_\text{aux} = 3$ layers in all experiments. The number of summary slots for each task is given in \Cref{tab:hyperparams}. The slots interact with the design--outcome tokens through self-attention, and the final mean keeps the main-encoder input length at one token per instance.

\paragraph{Reliability labels.}

For parameter-oriented BED, an instance token $\mathbf{e}_j = \mathbf{r}_k$ representing instance $\mathcal{A}_k$ receives the label $r_j^\text{inst}$ of \Cref{eq:label-aux}, thresholded at $1/2$ to give the binary label $c_j$. \Cref{tab:norm-ranges} lists the normalization ranges $[r_l^\text{min}, r_l^\text{max}]$ for CES and Psychometric, the two parameter-oriented benchmarks that use the instance channel, and \Cref{tab:hyperparams} gives their sharpness $\kappa$. For prediction-oriented BED, the regime-based binary labels are described in Appendix~\ref{app:exp-gp}.

\begin{table}[ht]
  \centering
  \small
  \caption{Normalization ranges $[r_l^\text{min}, r_l^\text{max}]$ used to construct reliability labels for previous instances.}
  \label{tab:norm-ranges}
  \begin{tabular}{llrr}
    \toprule
    Task & Parameter & $r_l^\text{min}$ & $r_l^\text{max}$ \\
    \midrule
    CES          & $\chi$                  & $0.05$ & $1$    \\
                 & $w_1$              & $0$    & $1$    \\
                 & $w_2$              & $0$    & $1$    \\
                 & $w_3$              & $0$    & $1$    \\
                 & $\log u$                & $-8$   & $8$    \\
    \midrule
    Psychometric & $\theta_\text{threshold}$    & $-3$   & $3$    \\
                 & $\theta_\text{slope}$    & $0.1$  & $2$    \\
                 & $\theta_\text{guess}$ & $0$    & $0.15$ \\
                 & $\theta_\text{lapse}$ & $0$    & $0.06$ \\
    \bottomrule
  \end{tabular}
\end{table}

\paragraph{Evaluation-time generation.}
We evaluate \method under two configurations that vary the number of instances $K$ while fixing their regime. \emph{$K$ relevant} draws all $K$ instances from the relevant regime, and \emph{$K$ irrelevant} draws all $K$ instances from the irrelevant regime, testing how performance scales with the amount of supplied information. Mixtures of relevant and irrelevant instances are evaluated on GP active learning, whose robustness sweep is described in Appendix~\ref{app:exp-gp}. The covariance $\Sigma_\text{rel}$ used at evaluation matches the training value.

\section{Experimental details}
\label{app:expdetails}

Unless stated otherwise, $\pm$ values and shaded bands denote 95\% confidence intervals of the mean over evaluation rollouts, computed as $\pm 1.96$ standard errors.

\subsection{BED tasks}
\label{app:bed-tasks}

\paragraph{Location Finding.}
\label{app:exp-location}
We adopt the source location task of \citet{sheng2005maximum} as formulated by \citet{foster2021deep}. The goal is to localize $N_{\mathrm{src}}$ hidden sources from intensity measurements at sensor locations $\mathbf{x} \in \mathbb{R}^d$. The expected signal at $\mathbf{x}$ follows an inverse-square attenuation law,
\begin{equation}
\mu(\boldsymbol{\theta}, \mathbf{x}) = b_{\mathrm{bg}} + \sum_{s=1}^{N_{\mathrm{src}}} \frac{\xi_s}{m + \lVert \boldsymbol{\theta}_s - \mathbf{x} \rVert^2},
\end{equation}
where $\xi_s > 0$ is the strength of source $s$, $b_{\mathrm{bg}} > 0$ is a background term, and $m > 0$ is a small damping constant. The observation is log-normal,
\begin{equation}
\log y \mid \boldsymbol{\theta}, \mathbf{x} \sim \mathcal{N}\bigl(\log \mu(\boldsymbol{\theta}, \mathbf{x}),\, \sigma^2\bigr).
\end{equation}
The prior on each source location $\boldsymbol{\theta}_s$ is uniform on $[0, 1]^d$. We use $N_{\mathrm{src}} = 2$, $d = 2$, $\xi_1 = \xi_2 = 1$, $b_{\mathrm{bg}} = 0.1$, $m = 10^{-4}$, and $\sigma = 0.5$.

\paragraph{Constant Elasticity of Substitution (CES).}
\label{app:exp-ces}
The CES task originates with \citet{arrow1961capital} and was adapted to the BED setting by \citet{foster2020unified}. The experimenter infers preference parameters by presenting pairs of bundles to a synthetic participant whose utility for a bundle of $N_{\mathrm{goods}}$ goods, $\mathbf{z} \in [0, 100]^{N_{\mathrm{goods}}}$, is
\begin{equation}
U(\mathbf{z}) = \Bigl(\sum_{i=1}^{N_{\mathrm{goods}}} w_i z_i^\chi\Bigr)^{1/\chi}.
\end{equation}
The latent parameters are $\boldsymbol{\theta} = (\chi, \boldsymbol{w}, u)$ with priors
\begin{align}
\chi &\sim \mathrm{Beta}(1, 1), \\
\boldsymbol{w} &\sim \mathrm{Dirichlet}(\mathbf{1}_{N_{\mathrm{goods}}}), \\
\log u &\sim \mathcal{N}(1, 3^2).
\end{align}
A design is a pair $\mathbf{x} = (\mathbf{z}, \mathbf{z}')$, and the response is generated by adding Gaussian noise to the utility difference and passing the result through a clipped sigmoid,
\begin{align}
\eta &\sim \mathcal{N}\Bigl(u\,(U(\mathbf{z}) - U(\mathbf{z}')),\; u^2 \sigma_{\mathrm{CES}}^2 (1 + \|\mathbf{z} - \mathbf{z}'\|)^2\Bigr), \\
y &= \mathrm{clip}\bigl(\sigma(\eta),\, \epsilon,\, 1 - \epsilon\bigr),
\end{align}
where $\sigma$ here denotes the logistic sigmoid. We use $N_{\mathrm{goods}} = 3$, $\sigma_{\mathrm{CES}} = 0.005$, and $\epsilon = 2^{-22}$, giving $\boldsymbol{\theta} \in \mathbb{R}^5$ and a six-dimensional design space.

\subsection{BED baselines}
\label{app:bed-baselines}
This subsection describes the baselines used on Location Finding and CES. For the random-design reference, we draw each $x_t$ uniformly from the design space.

\paragraph{Deep Adaptive Design (DAD).}
DAD \citep{foster2021deep} learns an amortized design policy by stochastic gradient ascent on the sequential Prior Contrastive Estimation (sPCE) lower bound. For a policy $\pi$ and $L_S \ge 0$ contrastive samples, the sPCE bound on the total information gain over $T$ steps is
\begin{equation}
\mathcal{L}_T(\pi, L_S) = \mathbb{E}_{p(\theta_0, \mathcal{D}_T \mid \pi)}\!\left[ \log \frac{p(\mathcal{D}_T \mid \theta_0, \pi)}{\frac{1}{L_S+1} \sum_{n=0}^{L_S} p(\mathcal{D}_T \mid \theta_n, \pi)} \right],
\end{equation}
where the contrastive samples $\theta_{1:L_S}$ are drawn independently from the prior $p(\theta)$. The DAD policy comprises a permutation-invariant encoder that aggregates the experimental history through pooling and an emitter that outputs the next design, following the architecture of \citet{foster2021deep}. Training uses Adam with an initial learning rate of $5 \times 10^{-5}$ and $(\beta_1,\beta_2) = (0.8, 0.998)$, together with an exponential schedule that decays the rate by a factor of $0.98$ every $1{,}000$ epochs.

\paragraph{RL-BOED.}
RL-BOED \citep{blau2022optimizing} casts the design problem as a Markov decision process and learns the design policy by reinforcement learning, using a stepwise reward that estimates the marginal contribution of the $t$-th step to the sequential EIG. The policy network shares its backbone with DAD but replaces the deterministic emitter by a Tanh-Gaussian distribution. The encoder is a two-layer MLP with $128$ ReLU units followed by a linear output of $64$ units. We train with Randomized Ensembled Double Q-learning (REDQ) \citep{chen2021randomized} using the configuration in \Cref{tab:rl-boed-hyperparams}.

\begin{table}[!ht]
\centering
\caption{Hyperparameters for RL-BOED \citep{blau2022optimizing} on the two BED tasks.}
\label{tab:rl-boed-hyperparams}
\begin{tabular}{lcc}
\toprule
\textbf{Parameter} & \textbf{Location Finding} & \textbf{CES} \\
\midrule
Critics & $2$ & $2$ \\
Random subsets & $2$ & $2$ \\
Training epochs & $10^5$ & $10^5$ \\
Discount factor $\gamma$ & $0.9$ & $0.9$ \\
Target update rate & $10^{-3}$ & $5 \times 10^{-3}$ \\
Policy learning rate & $10^{-4}$ & $3 \times 10^{-4}$ \\
Critic learning rate & $3 \times 10^{-4}$ & $3 \times 10^{-4}$ \\
Replay buffer size & $10^7$ & $10^6$ \\
\bottomrule
\end{tabular}
\end{table}

\paragraph{ALINE.}
ALINE \citep{huang2025aline} jointly trains the design policy and a posterior approximation network in a shared Transformer encoder. Tokens for the experimental history, candidate designs, and target slots are processed jointly, with attention masks that route information among the three groups. A design head produces a categorical distribution over the candidate designs and an inference head parameterizes a Gaussian-mixture posterior over the latent parameters. The policy is trained with REINFORCE \citep{williams1992simple} using a per-step reward defined as the one-step gain in posterior log-probability of the ground-truth target, and the inference network is trained by maximum likelihood. Training uses AdamW with learning rate $10^{-3}$ and weight decay $10^{-2}$. The remaining hyperparameters are summarized in \Cref{tab:aline-hyperparams}.

\begin{table}[!ht]
\centering
\caption{Hyperparameters for ALINE \citep{huang2025aline}.}
\label{tab:aline-hyperparams}
\scalebox{0.95}{
\begin{tabular}{lcccc}
\toprule
\textbf{Parameter} & \textbf{Location Finding} & \textbf{CES}  & \textbf{GP active learning} & \textbf{Psychometric} \\
\midrule
Transformer layers & $3$ & $3$ & $3$ & $3$\\
Attention heads & $4$ & $4$ & $4$ & $4$\\
GMM components & $10$ & $10$ & $10$ & $10$\\
Batch size & $200$ & $200$ & $200$ & $200$\\
Candidate pool size $N_q^\text{eval}$ & $2{,}000$ & $2{,}000$ & $500$ & $200$ \\
Total epochs $E$ & $10^{5}$ & $2 \times 10^{5}$ & $2.5 \times 10^5 $ & $1.2 \times 10^5$ \\
\bottomrule
\end{tabular}}
\end{table}

\paragraph{ALINE-aux.}
ALINE-aux shares \method's backbone and is retrained on the same $70/30$ relevant/irrelevant mixture of previous instances (\Cref{sec:filter}), receiving the $K \cdot T$ previous-instance observations by concatenation to the experimental history during both training and deployment. It therefore has the opportunity to learn to ignore unreliable instances on its own, and the gap between ALINE-aux and \method isolates the token interface and the filter from the effect of training exposure. 

\paragraph{Retrieve-then-condition.}
Retrieve-then-condition is a knowledge-selection baseline built on ALINE-aux. For each previous instance $\mathcal{A}_k$ we fit a single parameter estimate $\hat\theta^{(k)}$ by MAP estimation under the prior. At each round $t$ we score every instance by how well its estimate explains the data collected so far in the live experiment, $\sum_{(x, y) \in \mathcal{D}_t} \log p(y \mid \hat\theta^{(k)}, x)$, keep the two best-scoring instances, and pass them to ALINE-aux by concatenation.

\paragraph{Deployment-time concatenation.}
DAD\textsuperscript{*}, RL-BOED\textsuperscript{*}, and ALINE\textsuperscript{*} receive the $K$ previous instances without retraining by concatenating their $K \cdot T$ design--outcome pairs to the experimental history: as additional context tokens for ALINE\textsuperscript{*}, and as additional elements of the pooled history for the encoders of DAD\textsuperscript{*} and RL-BOED\textsuperscript{*}.

\subsection{Active learning for regression}
\label{app:exp-gp}

\begin{figure}[!ht]
\centering
\includegraphics[width=0.95\textwidth]{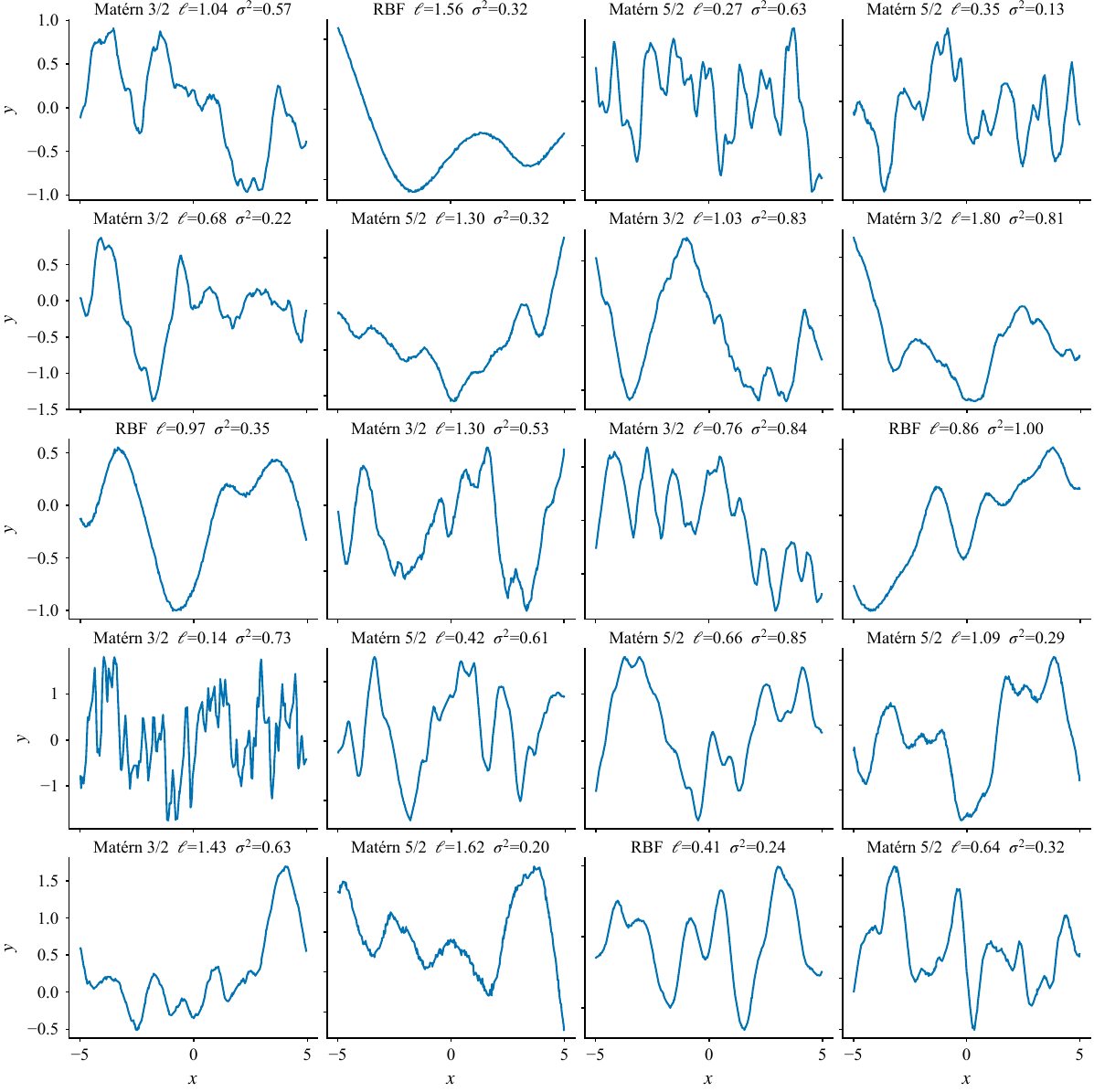}
\caption{Examples of synthetic 1D GP functions used to train \method, each drawn from the kernel and hyperparameter prior described above with noisy point evaluations.}
\label{fig:gp-train-samples}
\end{figure}

\paragraph{Synthetic training samples.}
We train \method on synthetic Gaussian process samples. Each training rollout draws a kernel uniformly from $\{\text{RBF},\,\text{Mat\'ern\,3/2},\,\text{Mat\'ern\,5/2}\}$; Mat\'ern\,1/2 is excluded from the training family because its sample paths are too rough to be informative for our active learning targets. The output scale is sampled from $\mathcal{U}(0.1, 1)$ and the lengthscale from $\mathcal{U}(0.1, 2)$. Inputs are drawn uniformly from $[-5, 5]$, and outputs are perturbed by Gaussian noise with standard deviation $\sigma_\text{train} = 0.01$. Examples of training samples are shown in \Cref{fig:gp-train-samples}.

\begin{figure}[!ht]
\centering
\includegraphics[width=0.95\textwidth]{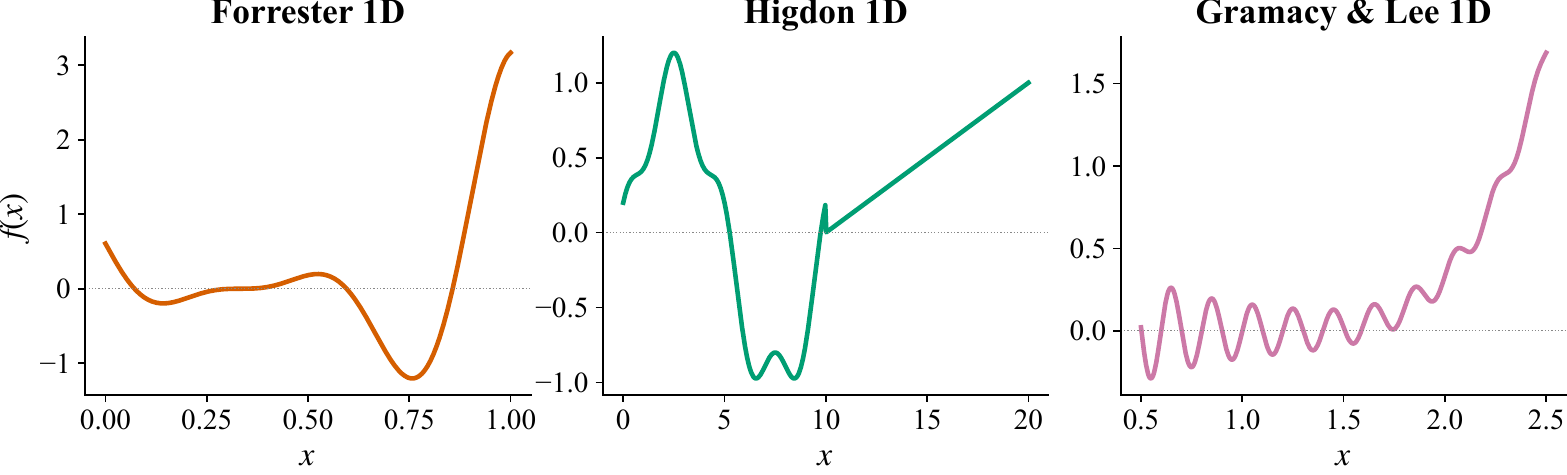}
\caption{The three analytic 1D benchmark functions with the output scaling used at evaluation. The horizontal axes retain their native input coordinates.}
\label{fig:gp-benchmark}
\end{figure}

\paragraph{Benchmark functions.}
We evaluate on four 1D benchmarks: Synthetic, Forrester, Higdon, and Gramacy \& Lee. Synthetic uses fresh draws from the training GP prior on $[-5, 5]$. The three analytic functions in \Cref{fig:gp-benchmark} have native input domains $[0, 1]$, $[0, 20]$, and $[0.5, 2.5]$, respectively. For model inputs, each native domain is affinely mapped to $x_\text{norm} \in [-5, 5]$, matching the training input range. To reduce differences in output magnitude across benchmarks, we divide Forrester outputs by $5$ and Gramacy \& Lee outputs by $3$; RMSE is reported on the resulting output scale.

\begin{figure}[!ht]
\centering
\includegraphics[width=0.92\textwidth]{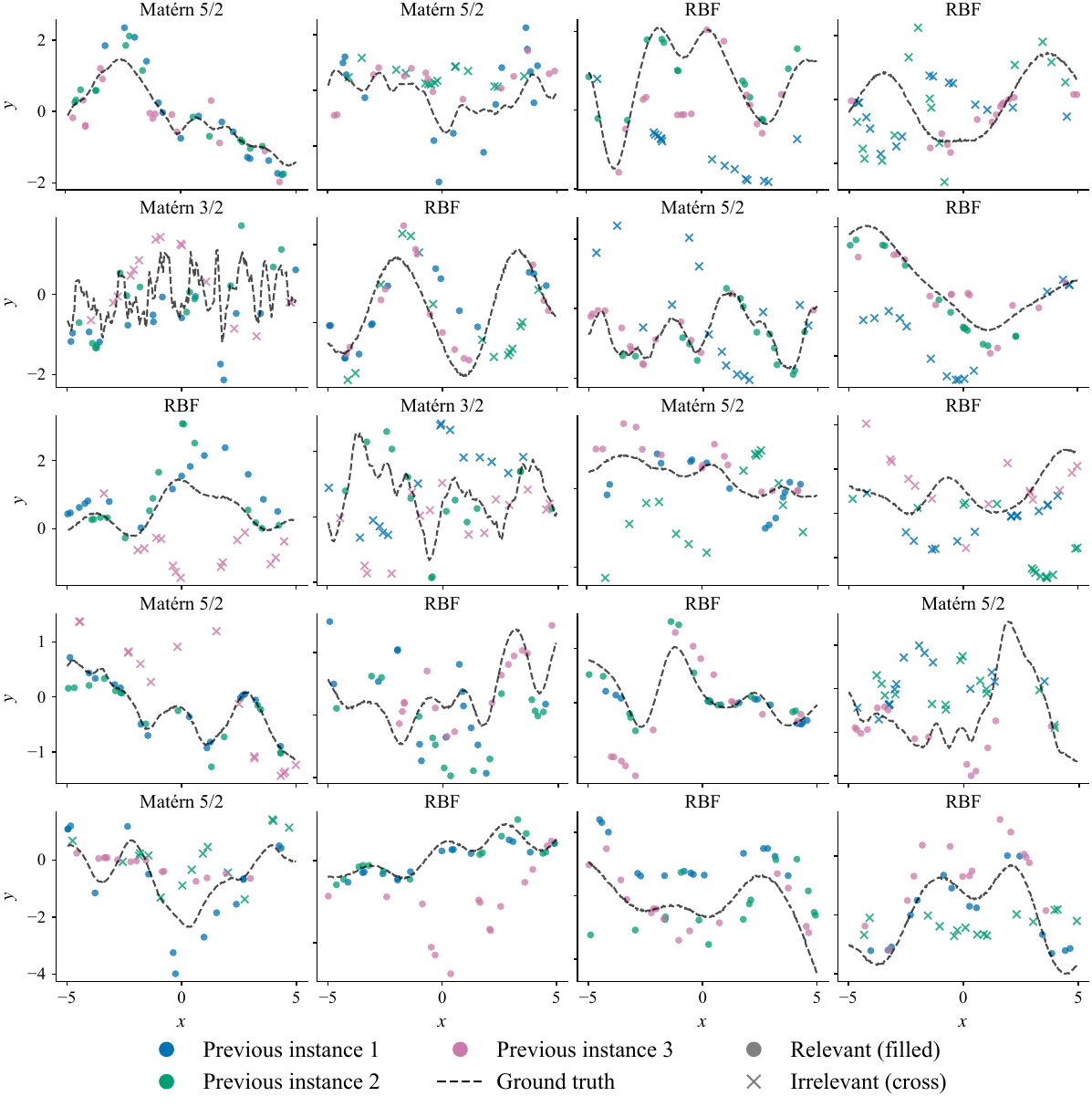}
\caption{Examples of previous instances supplied during training. Each panel shows one target function (black dashed line) and up to three instances $\mathcal{A}_1, \mathcal{A}_2, \mathcal{A}_3$ (colored points), with filled circles for the relevant regime and crosses for the irrelevant regime.}
\label{fig:gp-aux-samples}
\end{figure}

\paragraph{Instance generation.}
Each rollout supplies $K$ previous instances, of which $K_\text{rel}$ are drawn from the relevant regime and $K - K_\text{rel}$ from the irrelevant regime. During training $K \sim \mathrm{Unif}\{0, 1, \ldots, 5\}$; in the robustness sweep of \Cref{sec:exp-gp}, $K = 5$ is fixed and $K_\text{rel}$ is swept from $0$ to $5$. All instances share $T$ design points $x_1, \ldots, x_T$ drawn uniformly in $x_\text{norm}$; their outputs are generated as follows, and examples are shown in \Cref{fig:gp-aux-samples}.

\textit{Relevant regime.}
On the three analytic benchmarks, a relevant instance is a per-instance affine perturbation of the target function,
\begin{equation}
y_i^{(k)} = a_k\, f^\star(x_i) + b_k + \varepsilon_i^{(k)}, \qquad i = 1, \ldots, T,
\label{eq:gp-rel-affine}
\end{equation}
with $a_k \sim \mathcal{U}(0.5, 1.5)$, $b_k \sim \mathcal{U}(-0.5, 0.5) \cdot \sigma_y$ where $\sigma_y$ is the standard deviation of $f^\star$ over the input domain, and $\varepsilon_i^{(k)} \sim \mathcal{N}(0, \sigma_\text{train}^2)$. The per-instance scale and offset ensure that simply pooling the instances with the target data does not recover $f^\star$ exactly. On the synthetic benchmark, where $f^\star$ is itself a GP draw, the target and the instance are drawn jointly from the same GP (same kernel and hyperparameters, noise level $\sigma_\text{train}$), and the resulting instance observations $y^{(k)}_\text{base}$ are perturbed by an independent GP function $f_\text{bias}^{(k)}$ drawn from the training prior, with scale $s_k \sim \mathcal{U}(0.2, 2)$ and mode $m_k \sim \mathrm{Bern}(0.5)$:
\begin{equation}
y_i^{(k)} =
\begin{cases}
y_{\text{base},i}^{(k)} + s_k\, f_\text{bias}^{(k)}(x_i) & \text{if } m_k = 0 \ \text{(additive)}, \\[2pt]
y_{\text{base},i}^{(k)} \bigl(1 + s_k \tanh f_\text{bias}^{(k)}(x_i)\bigr) & \text{if } m_k = 1 \ \text{(multiplicative)}.
\end{cases}
\label{eq:gp-rel-synth}
\end{equation}

\textit{Irrelevant regime.}
An irrelevant instance is an independent GP draw whose hyperparameters are drawn from the training prior $p(\theta)$,
\begin{equation}
\theta^{(k)} \sim p(\theta), \qquad y_i^{(k)} \sim p(y \mid \theta^{(k)}, x_i).
\label{eq:gp-irrel}
\end{equation}
On the synthetic benchmark this is implemented as a batch permutation of the target hyperparameters.

\textit{Reliability label.}
Because the task is non-parametric, the parameter-distance label of \Cref{eq:label-aux} does not apply. For the token $\mathbf{e}_j = \mathbf{r}_k$ representing instance $k$, we instead supervise the filter with the regime indicator,
\begin{equation}
c_j =
\begin{cases}
1 & \text{if instance $k$ is drawn from the relevant regime (\Cref{eq:gp-rel-affine,eq:gp-rel-synth}),} \\
0 & \text{if it is drawn from the irrelevant regime (\Cref{eq:gp-irrel}).}
\end{cases}
\label{eq:gp-rel-label}
\end{equation}

\paragraph{Baseline implementations.}
The three baselines on this task ingest previous-instance observations without the token interface, and none of them uses the reliability labels.

\textit{ALINE\textsuperscript{*}.} At deployment, the ALINE backbone receives all $K \cdot T$ previous-instance observations as additional context tokens, concatenated with the experimental observations. This is the same concatenation protocol used for ALINE\textsuperscript{*} on the CES task (\Cref{sec:exp-ces}).

\textit{GP\textsuperscript{*}.} A Gaussian process surrogate conditioned on the experimental observations and all $K \cdot T$ previous-instance observations together, treating every point as an evaluation of the same function. The GP uses an RBF kernel with a noise term, is refit by marginal-likelihood maximization at each step, and selects queries by uncertainty sampling. This is the variant reported in \Cref{sec:exp-gp}. 

\textit{MT-GP.} A multi-task Gaussian process that treats each previous instance as a separate task correlated with the target. The target is task $0$ and each auxiliary previous instance is task $k \in \{1, \ldots, K\}$. We use the rank-$Q$ intrinsic coregionalization model of \citet{bonilla2007multi} with a zero prior mean,
\begin{equation}
k\bigl((x_i, t_i),\, (x_j, t_j)\bigr)
= \bigl[W W^{\top} + \mathrm{diag}(\boldsymbol{v})\bigr]_{t_i, t_j}
\,\cdot\,
\sigma_f^{2}\, \exp\!\Bigl(-\tfrac{1}{2}\,\|x_i - x_j\|^{2} / \ell^{2}\Bigr),
\label{eq:app-mtgp-kernel}
\end{equation}
where the tasks share an RBF kernel with lengthscale $\ell$ and amplitude $\sigma_f^2$, while $W$ and $\boldsymbol{v}$ determine the correlations and task-specific variation across instances. We use rank $Q=1$, so tasks share a single latent component together with task-specific residual variation. Each task additionally has its own learned observation-noise variance. Hyperparameters are optimized by maximum marginal likelihood using Adam with learning rate $0.05$, with $200$ optimization steps for the initial fit and $60$ steps for subsequent fits initialized from the preceding solution. At each acquisition step, the GP is fit to the currently observed target points together with all previous-instance observations, and the next target point is selected by maximizing the predictive standard deviation for task $0$.

\paragraph{Acquisition functions.}
Random sampling selects the next query from the candidate pool randomly.

Uncertainty sampling selects the next query at the point of maximum predictive standard deviation,
\begin{equation}
\mathrm{US}(x) = \sqrt{\mathbb{V}[y \mid x, \mathcal{D}]}.
\label{eq:app-us}
\end{equation}
Variance reduction selects the query that maximizes the expected reduction in posterior variance over a test set $\{x_m^\star\}_{m=1}^M$ \citep{yu2006active},
\begin{equation}
\mathrm{VR}(x) = \frac{\sum_{m=1}^{M} \bigl(\mathrm{Cov}_\text{post}(x_m^\star, x)\bigr)^2}
                      {\mathbb{V}[y \mid x, \mathcal{D}]},
\label{eq:app-vr}
\end{equation}
with the posterior covariance computed in closed form from the GP,
\begin{equation}
\mathrm{Cov}_\text{post}(x^\star, x) = k(x^\star, x)
- k(x^\star, X_\text{train})\,(K_\text{train} + \sigma_{\mathrm{GP}}^2 I)^{-1}\, k(X_\text{train}, x),
\label{eq:app-cov-post}
\end{equation}
where $k$ is the GP kernel, $K_\text{train} = k(X_\text{train}, X_\text{train})$, and $\sigma_{\mathrm{GP}}^2$ is the noise variance.

\subsection{Psychometric model}
\label{app:experiment-psy}

\paragraph{Model description.}
\label{app:psychometric-details}
In the psychometric task \citep{watson2017quest+}, the experimenter infers the parameters of a participant's psychometric function from binary yes/no responses to stimulus intensities $x \in \mathbb{R}$. The response probability is
\begin{equation}
\pi(x; \boldsymbol{\theta}) = \theta_\text{guess} + (1 - \theta_\text{guess} - \theta_\text{lapse})\, F\!\left(\frac{x - \theta_\text{threshold}}{\theta_\text{slope}}\right), \qquad F(z) = 1 - \exp(-10^{z}),
\label{eq:psy-model}
\end{equation}
where $\theta_\text{threshold}$ is the threshold, $\theta_\text{slope}$ the slope, $\theta_\text{guess}$ the guess rate, and $\theta_\text{lapse}$ the lapse rate. The priors are
\begin{align}
\theta_\text{threshold} &\sim \mathcal{U}(-3, 3), & \theta_\text{slope} &\sim \mathcal{U}(0.1, 2), \\
\theta_\text{guess} &\sim \mathcal{U}(0, 0.15), & \theta_\text{lapse} &\sim \mathcal{U}(0, 0.06),
\end{align}
where the narrow supports on $\theta_\text{guess}$ and $\theta_\text{lapse}$ reflect their physiological interpretation. The design space is $[-5, 5]$ and the response is $y \sim \mathrm{Bernoulli}(\pi(x; \boldsymbol{\theta}))$. The metric is the joint RMSE on $(\theta_\text{threshold}, \theta_\text{slope})$.

\paragraph{Baselines.}
We compare \method against three baselines.

\textit{QUEST+} \citep{watson2017quest+} selects each stimulus to maximize the expected information gain about the full parameter vector on a discrete grid of candidate values. We discretize the parameter space at the resolution recommended by \citet{watson2017quest+} and run inference over the joint posterior on $(\theta_\text{threshold}, \theta_\text{slope}, \theta_\text{guess}, \theta_\text{lapse})$, reporting RMSE on $(\theta_\text{threshold}, \theta_\text{slope})$.

\textit{Psi-marginal} \citep{prins2013psi} extends the original psi method \citep{kontsevich1999bayesian} by marginalizing over nuisance parameters during stimulus selection. Since our metric targets only $(\theta_\text{threshold}, \theta_\text{slope})$, we configure Psi-marginal to optimize the information gain on these two parameters and treat $(\theta_\text{guess}, \theta_\text{lapse})$ as nuisance parameters.

\textit{ALINE} \citep{huang2025aline} is configured with a parameter target $Z = (\theta_\text{threshold}, \theta_\text{slope}, \theta_\text{guess}, \theta_\text{lapse})$ in line with our setting. The architecture and training match the description in Appendix~\ref{app:bed-baselines}.

\section{Additional experimental results}
\label{app:addexpdetails}

\subsection{Location Finding}
\label{app:location-trajectories}

\paragraph{Design trajectory.}

\begin{figure}[!ht]
\centering
\includegraphics[width=0.55\textwidth]{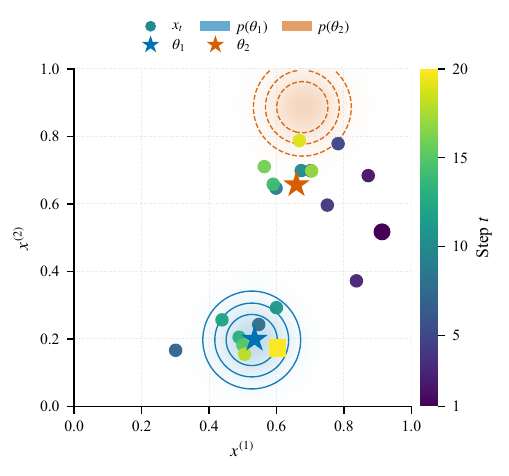}
\caption{Location Finding trajectory: prior contours, true sources (stars), and queries colored by step; the yellow square marks the final query.}
\label{fig:lf-trajectory}
\end{figure}

In \Cref{fig:lf-trajectory}, most queries lie in or near the regions highlighted by the supplied prior, rather than being spread across the full design space.

\begin{figure}[!ht]
\centering
\includegraphics[width=0.65\textwidth]{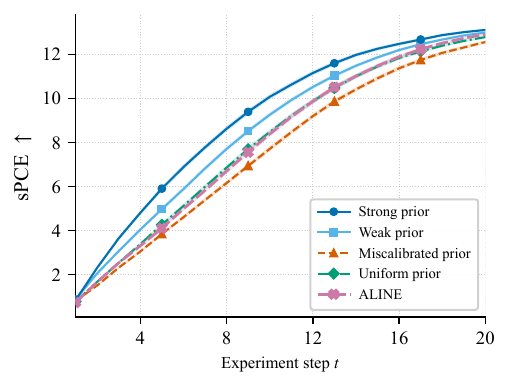}
\caption{sPCE ($\uparrow$) on Location Finding: \method under strong, weak, uniform, and miscalibrated priors, and baseline ALINE (mean $\pm$ 95\% CI over $2{,}000$ rollouts).}
\label{fig:lf-prior-comparison}
\end{figure}

\paragraph{Effect of prior type.}
The main-text result in \Cref{fig:location} shows ALINE outperforming Random, DAD, and RL-BOED, so we restrict the prior-type comparison here to ALINE and \method. \Cref{fig:lf-prior-comparison} reports sPCE under four prior conditions (strong, weak, miscalibrated, and uniform) alongside the baseline (ALINE). \method accelerates early-stage information acquisition under a strong prior and to a smaller extent under a weak prior, and keeps close to the no-external baseline (ALINE) under both miscalibrated and uniform priors.

\paragraph{Filter outputs over time.}

\begin{figure}[!ht]
\centering
\includegraphics[width=0.65\textwidth]{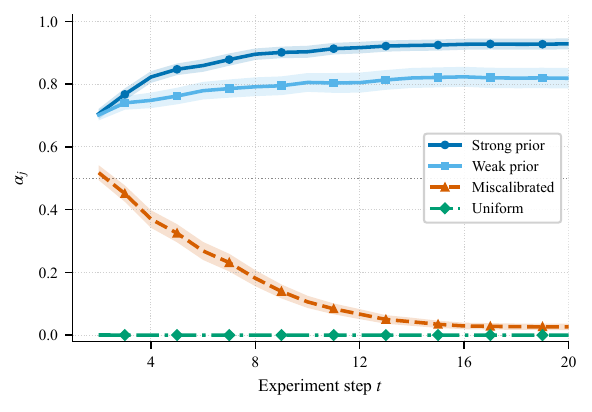}
\caption{Reliability filter output $\alpha_j$ for the prior tokens on Location Finding, mean $\pm$ 95\% CI over $100$ rollouts, shown as a function of experiment step $t$. Each curve corresponds to one of the four prior conditions in \Cref{fig:lf-prior-comparison}. The dotted line marks $\tau=0.5$.}
\label{fig:lf-alpha}
\end{figure}

\Cref{fig:lf-alpha} traces the per-token reliability filter outputs $\alpha_j$ over $T = 20$ rounds under the four prior conditions. Strong-prior scores settle close to $1$, while weak-prior scores stabilize around $0.8$. Miscalibrated-prior scores start at an intermediate level and decay toward $0$ as observations accumulate, indicating decreasing trust in the provided prior. Uniform-prior scores stay near $0$ from the first rounds. These patterns match the supervision signal described in Appendix~\ref{app:prior}.

\subsection{CES}
\label{app:ces_results}

\paragraph{Posterior visualization.}

\begin{figure}[!ht]
\centering
\includegraphics[width=1.0\textwidth]{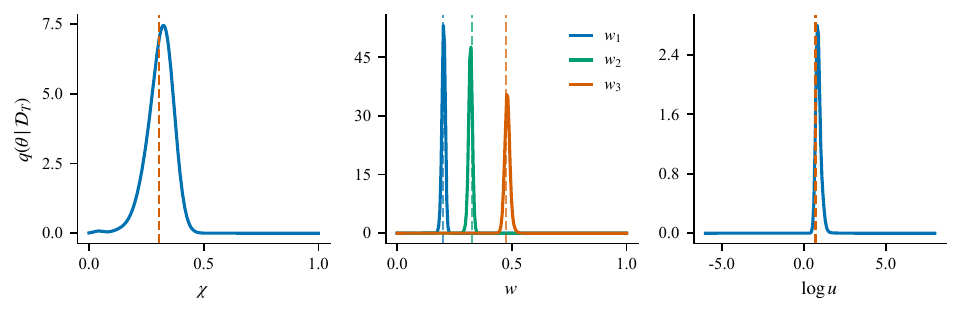}
\caption{Marginal posterior distributions estimated by \method on the CES task after $T = 10$ steps. Dashed lines mark the true parameter values.}
\label{fig:ces-posterior}
\end{figure}

\Cref{fig:ces-posterior} shows the marginal posterior distributions estimated by \method on a single CES rollout. The posteriors concentrate around the ground-truth values across all five parameters.

\begin{figure}[!ht]
\centering
\includegraphics[width=1.0\textwidth]{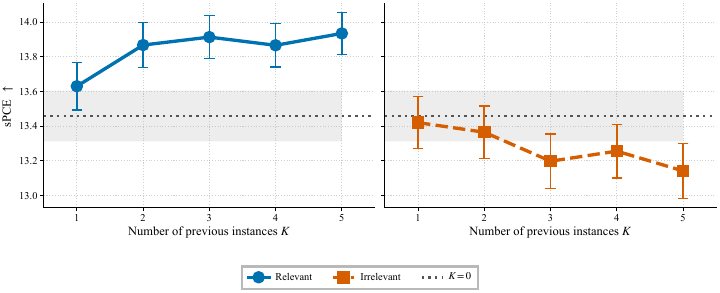}
\caption{Effect of the number of previous instances $K$ on CES. sPCE (mean $\pm$ 95\% CI, $2{,}000$ rollouts).}
\label{fig:ces-aux-sweep}
\end{figure}

\paragraph{Effect of the number of previous instances $K$.}
\Cref{fig:ces-aux-sweep} contrasts the effect of adding relevant and irrelevant previous instances. With relevant instances (left), sPCE improves initially and then levels off, suggesting diminishing returns from additional context. With irrelevant instances (right), sPCE remains close to its $K=0$ value over the tested range.

\paragraph{Effect of training-time exposure to previous instances.}

\begin{figure}[!ht]
\centering
\includegraphics[width=1.0\textwidth]{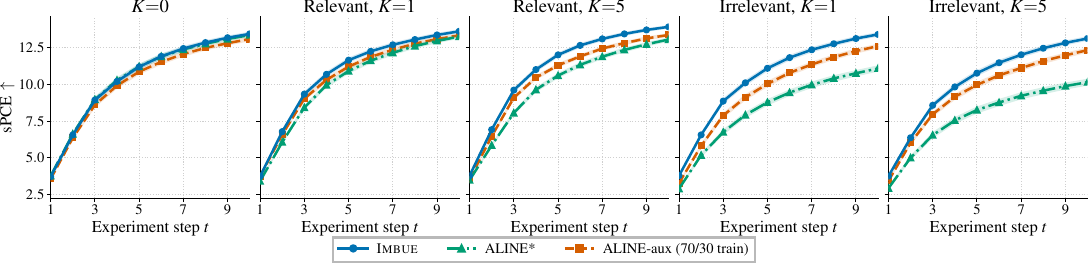}
\caption{CES with previous-instance observations. Stepwise sPCE (mean $\pm$ 95\% CI, $2{,}000$ rollouts) over experiment steps $t = 1, \ldots, 10$. Panels vary the number of previous instances supplied at deployment, $K = 0$, $K \in \{1, 5\}$ relevant, or $K \in \{1, 5\}$ irrelevant.}
\label{fig:ces-aline-aux-train}
\end{figure}

\Cref{fig:ces-aline-aux-train} extends \Cref{fig:motiv} to the five configurations in \Cref{tab:ces-results}. With auxiliary inputs, ALINE-aux improves on ALINE\textsuperscript{*}, especially under irrelevant input, but remains below \method. This gap supports the combined benefit of the instance-token interface and reliability filter beyond training exposure. Their individual contributions are examined in \Cref{tab:ces-ablation}. At $K = 0$, the three methods perform similarly.

\paragraph{Number of summary slots.}
We ablate the number of summary slots $n_s$ in the per-instance encoder $g_\text{aux}$ on the CES task. \Cref{tab:summary-slots} shows small differences across these settings. Increasing the number of summary slots does not consistently improve performance. We use $n_s = 2$ in the CES experiments and $n_s = 4$ in the GP and Psychometric experiments.

\begin{table}[!ht]
\centering
\caption{Effect of the number of summary slots $n_s$ in the per-instance encoder $g_\text{aux}$ on CES. sPCE (mean $\pm$ 95\% CI, $2{,}000$ rollouts).}
\label{tab:summary-slots}
\setlength{\tabcolsep}{5pt}
\begin{tabular}{l c c c c c}
\toprule
& $K = 0$ & \multicolumn{2}{c}{$K = 1$} & \multicolumn{2}{c}{$K = 5$} \\
\cmidrule(lr){2-2} \cmidrule(lr){3-4} \cmidrule(lr){5-6}
$n_s$ & --- & Rel. & Irrel. & Rel. & Irrel. \\
\midrule
$1$ & $13.42 \pm 0.15$ & $\mathbf{13.73 \pm 0.13}$ & $13.37 \pm 0.16$ & $13.92 \pm 0.12$ & $\mathbf{13.22 \pm 0.16}$ \\
$2$ (\textbf{\method}) & $\mathbf{13.46 \pm 0.15}$ & $13.63 \pm 0.14$ & $\mathbf{13.42 \pm 0.15}$ & $\mathbf{13.93 \pm 0.12}$ & $13.14 \pm 0.16$ \\
$4$ & $13.41 \pm 0.15$ & $13.58 \pm 0.14$ & $13.33 \pm 0.16$ & $13.82 \pm 0.12$ & $13.09 \pm 0.16$ \\
\bottomrule
\end{tabular}
\end{table}

\subsection{GP active learning}
\label{app:gp_results}

\paragraph{Acquisition trajectory.}

\begin{figure}[!ht]
\centering
\includegraphics[width=1.0\textwidth]{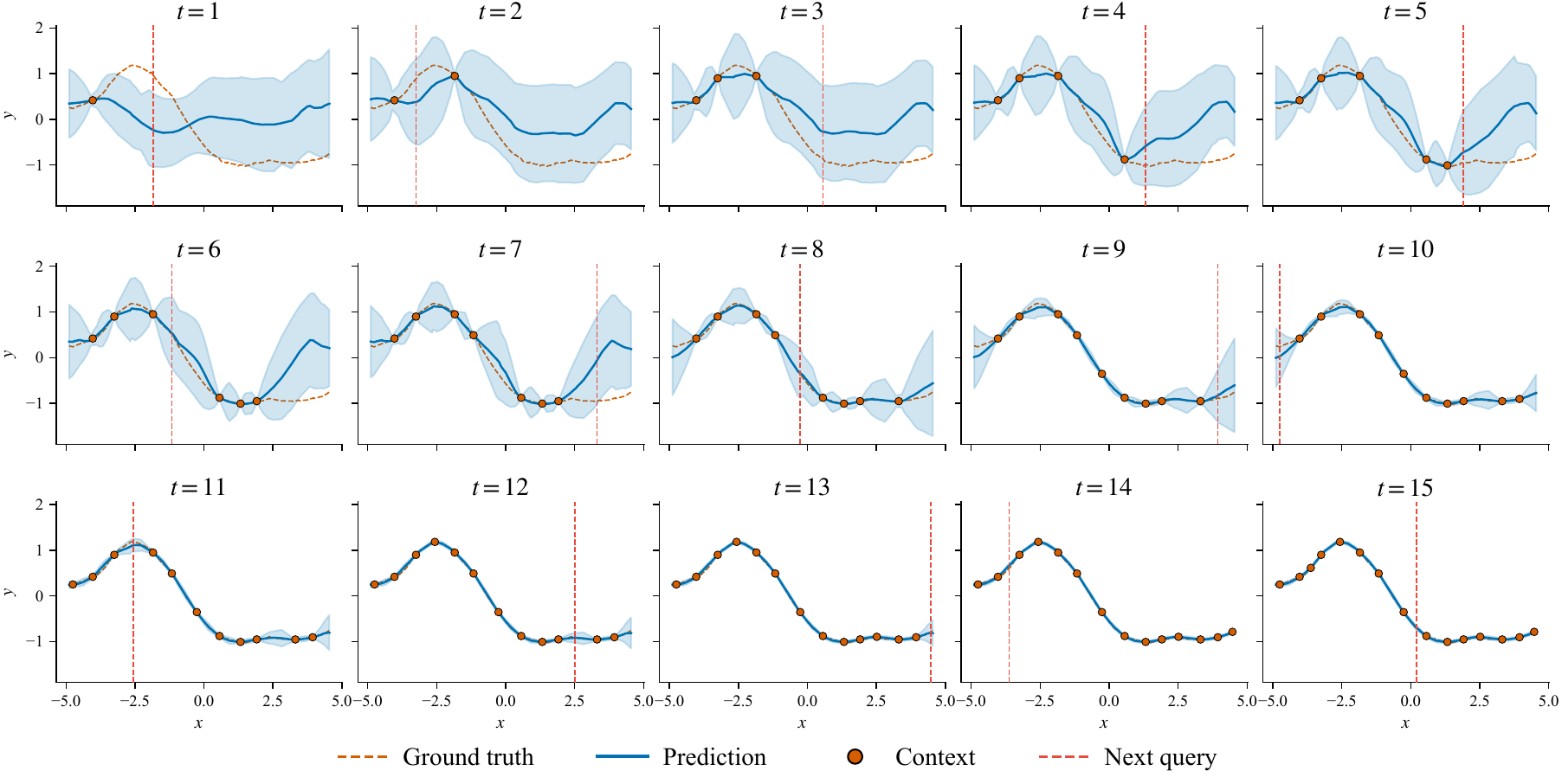}
\caption{Sequential design strategy of \method on a 1D synthetic GP function over $T = 15$ steps. Each panel shows the ground-truth function (orange dashed), the predictive mean (blue solid) with $95\%$ predictive band (shaded), the accumulated context (orange dots), and the next queried design (red dashed vertical line).}
\label{fig:gp-trajectory}
\end{figure}

\Cref{fig:gp-trajectory} shows one \method rollout on a synthetic GP function over $T = 15$ steps. As observations accumulate, the predictive mean approaches the ground truth and uncertainty decreases near queried locations.

\paragraph{Extended baseline comparison.}

\begin{figure}[!ht]
\centering
\includegraphics[width=1.0\textwidth]{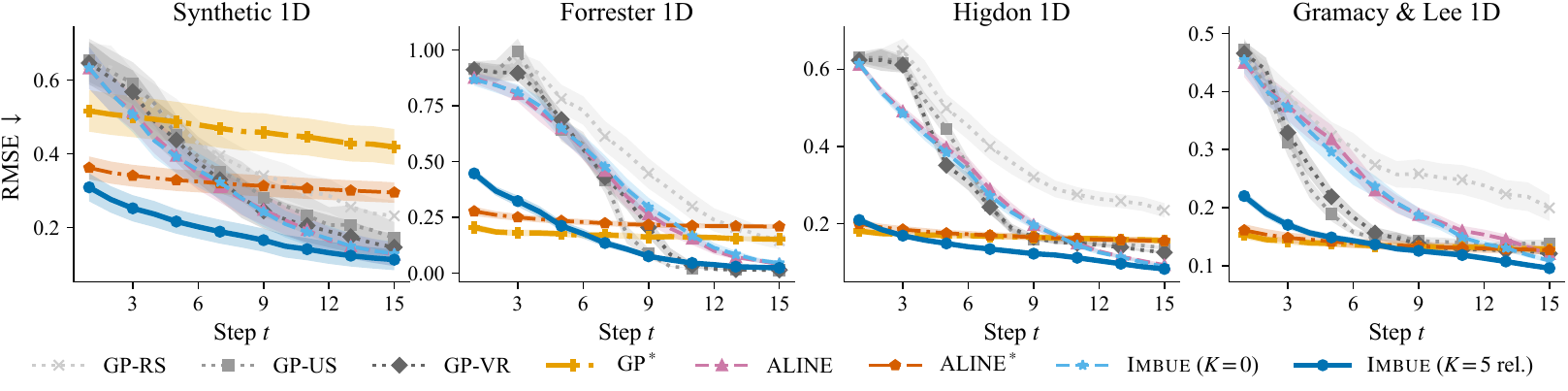}
\caption{Extended baseline comparison on four 1D GP active learning benchmarks (RMSE $\downarrow$, mean and 95\% CI over 100 rollouts).}
\label{fig:gp-baselines-extended}
\end{figure}

\Cref{fig:gp-baselines-extended} compares \method with GP and ALINE baselines. GP-RS, GP-US, and GP-VR condition on experimental observations and use random sampling, uncertainty sampling, and variance reduction \citep{yu2006active}, respectively. GP\textsuperscript{*} additionally conditions on all observations from five relevant previous instances and uses uncertainty sampling. ALINE uses experimental observations alone, while ALINE\textsuperscript{*} concatenates previous-instance observations at deployment. \method uses no previous instances ($K=0$) or five relevant instances ($K = 5$ rel.).

Without previous instances, \method performs similarly to ALINE and is competitive with the GP baselines. Supplying relevant previous instances reduces its early-step RMSE on all four benchmarks, with smaller differences from the $K=0$ variant later in the rollout. GP\textsuperscript{*} and ALINE\textsuperscript{*} have lower initial RMSE than their counterparts without previous instances, but improve less as new observations are collected. With relevant previous instances, \method attains lower final RMSE than both baselines.

\begin{figure}[!ht]
\centering
\includegraphics[width=1.0\textwidth]{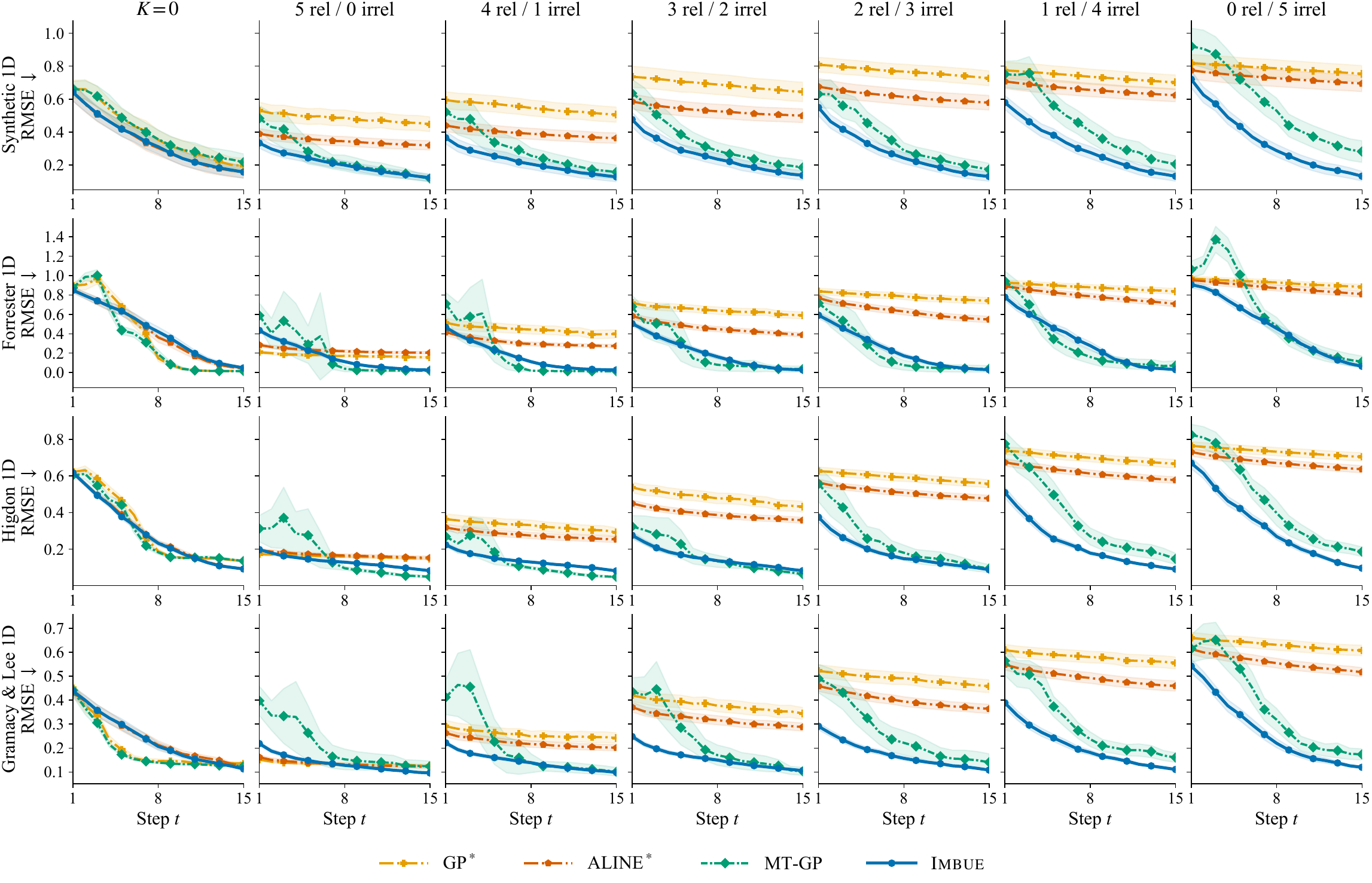}
\caption{RMSE at each experimental step on four 1D GP benchmarks ($\downarrow$, mean and 95\% CI over $100$ rollouts). Rows index the benchmark, and columns show the composition of the $K = 5$ previous instances as $\text{relevant}/\text{irrelevant}$ (the leftmost column is the $K = 0$ baseline).}
\label{fig:gp-robustness-full}
\end{figure}

\paragraph{Robustness to irrelevant previous instances.}
\Cref{fig:gp-robustness-full} reports RMSE at each experimental step for different proportions of irrelevant previous instances, extending the final-step comparison in the main text (\Cref{fig:gp_comparison}). When all five previous instances are relevant, \method starts with lower RMSE than in the $K=0$ condition on all four benchmarks. When irrelevant instances are included, \method continues to reduce RMSE as experimental observations accumulate, reaching final RMSE close to its $K=0$ performance across the tested compositions. In comparison, GP\textsuperscript{*} and ALINE\textsuperscript{*} show smaller reductions in RMSE over the rollout, and their final RMSE increases with the proportion of irrelevant instances.

\paragraph{Inference time.}

\Cref{tab:gp-time} reports total inference time for rollouts of $15$ steps on the four GP benchmarks. At $K=0$, \method has inference time comparable to ALINE. With five relevant previous instances, \method remains substantially faster than the GP variants in the table, which refit their models at each step. Relative to ALINE, \method incurs a small absolute increase in inference time for instance encoding and reliability filtering.

\begin{table}[!ht]
\centering
\caption{Total inference time for rollouts of $15$ steps on four GP benchmarks (mean $\pm$ standard deviation in seconds, $10$ runs).}
\label{tab:gp-time}
\setlength{\tabcolsep}{6pt}
\begin{tabular}{l c c c c}
\toprule
Method & Synthetic 1D & Forrester & Higdon & Gramacy \& Lee \\
\midrule
GP-RS & $0.332 \pm 0.020$ & $0.343 \pm 0.029$ & $0.305 \pm 0.024$ & $0.300 \pm 0.022$ \\
GP-US & $0.408 \pm 0.024$ & $0.381 \pm 0.020$ & $0.397 \pm 0.023$ & $0.399 \pm 0.027$ \\
GP-VR & $1.236 \pm 0.037$ & $1.242 \pm 0.020$ & $1.240 \pm 0.027$ & $1.216 \pm 0.033$ \\
GP\textsuperscript{*} ($K=5$ rel.) & $0.556 \pm 0.068$ & $0.640 \pm 0.036$ & $0.648 \pm 0.050$ & $0.649 \pm 0.087$ \\
ALINE & $0.047 \pm 0.000$ & $0.049 \pm 0.003$ & $0.049 \pm 0.003$ & $0.048 \pm 0.003$ \\
\midrule
\textbf{\method ($K=0$)} & $0.049 \pm 0.004$ & $0.047 \pm 0.000$ & $0.047 \pm 0.000$ & $0.047 \pm 0.000$ \\
\textbf{\method ($K = 5$ rel.)} & $0.074 \pm 0.000$ & $0.075 \pm 0.005$ & $0.075 \pm 0.004$ & $0.075 \pm 0.005$ \\
\bottomrule
\end{tabular}
\end{table}

\subsection{Psychometric model}
\label{app:psychometric_results}

\begin{figure}[!ht]
\centering
\includegraphics[width=1.0\textwidth]{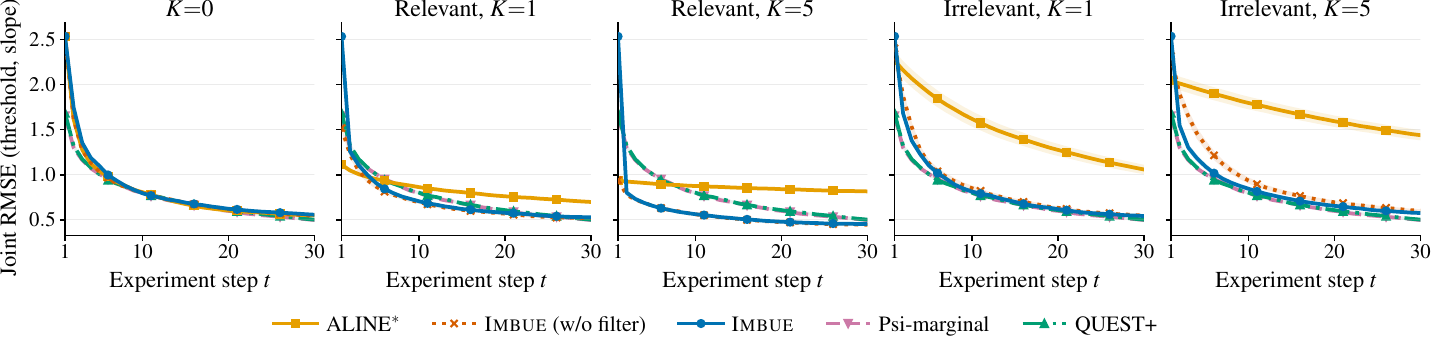}
\caption{Psychometric model: joint RMSE on $(\theta_\text{threshold}, \theta_\text{slope})$ ($\downarrow$) over $T = 30$ steps (mean $\pm$ 95\% CI, $500$ rollouts). Panels compare no instances ($K = 0$) with $K \in \{1, 5\}$ relevant or irrelevant instances. QUEST+ and Psi-marginal use no previous instances; their curves are repeated across panels for reference. The \method (w/o filter) curves are discussed in Appendix~\ref{app:ablation-and-sensitivity}.}
\label{fig:psy-aux-ablation}
\end{figure}

\Cref{fig:psy-aux-ablation} extends the main-text comparison (\Cref{fig:psy}) to different numbers of relevant or irrelevant previous instances. Without previous instances ($K=0$), \method performs similarly to ALINE, QUEST+, and Psi-marginal. Relevant previous instances accelerate the reduction in \method's RMSE, with greater benefits when more instances are supplied. ALINE\textsuperscript{*}, however, does not improve over its $K=0$ baseline even with relevant previous instances. With irrelevant previous instances, \method's RMSE remains close to that obtained with $K=0$ throughout the rollout, whereas the RMSE of ALINE\textsuperscript{*} increases with $K$.

\subsection{Ablations and sensitivity studies}
\label{app:ablation-and-sensitivity}

We examine the reliability filter across input channels, compare binary and soft gating, and assess sensitivity to the training mixture and filter loss weight.

\paragraph{Filter behavior across channels and reliability regimes.}

\begin{figure}[!ht]
\centering
\begin{subfigure}{0.95\textwidth}
\includegraphics[width=\textwidth]{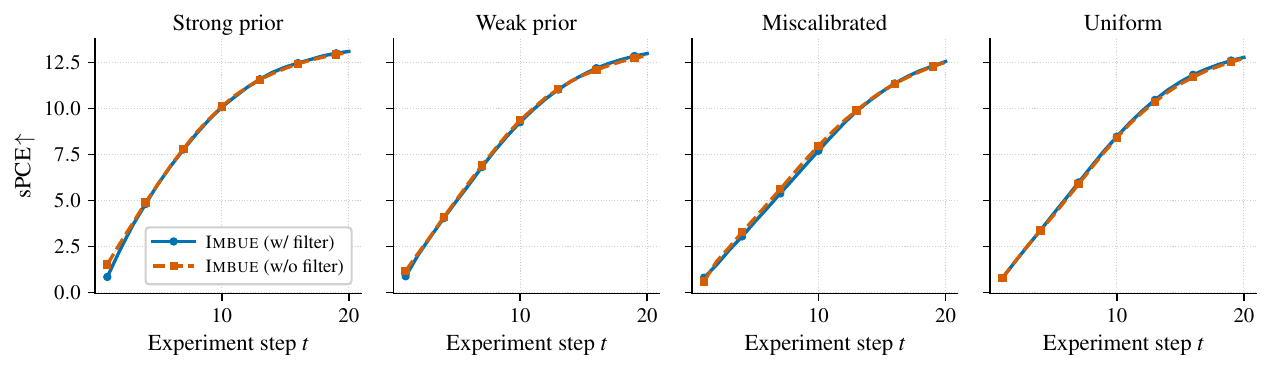}
\caption{Location Finding under four prior conditions.}
\label{fig:filter-ablation-lf}
\end{subfigure}\\[1em]
\begin{subfigure}{1.0\textwidth}
\includegraphics[width=\textwidth]{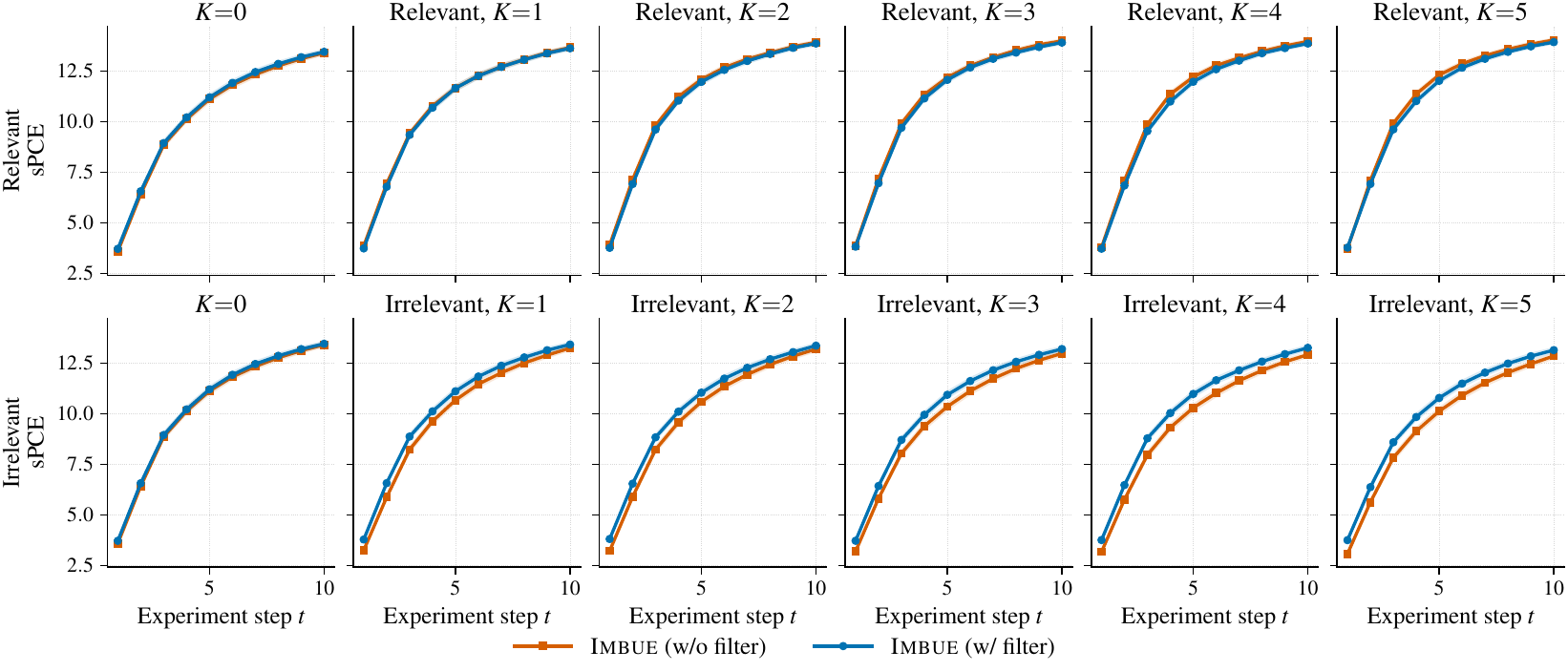}
\caption{CES under $K \in \{0, 1, \ldots, 5\}$ previous instances in both reliability regimes.}
\label{fig:filter-ablation-ces}
\end{subfigure}
\caption{Filter ablation on Location Finding and CES. sPCE mean $\pm$ 95\% CI over $2{,}000$ rollouts for the full \method (w/ filter) and the variant retrained without reliability filter (w/o filter).}
\label{fig:filter-ablation-all}
\end{figure}

\Cref{fig:filter-ablation-all} compares \method with and without the reliability filter on Location Finding (prior input) and CES (previous-instance input), with corresponding psychometric curves shown in \Cref{fig:psy-aux-ablation} as \method (w/o filter).

\textit{Prior channel.} On Location Finding (\Cref{fig:filter-ablation-lf}), the two variants are essentially indistinguishable across all four prior conditions, including the miscalibrated one.

\textit{Previous-instance channel under reliable input.} On CES with $K$ relevant instances (top row of \Cref{fig:filter-ablation-ces}) and on the psychometric task with $K \in \{1, 5\}$ relevant instances (relevant panels of \Cref{fig:psy-aux-ablation}), the two variants are again indistinguishable. When all instance tokens carry useful signal, removing the filter neither helps nor hurts.

\textit{Previous-instance channel under unreliable input.} On CES with irrelevant previous instances (bottom row of \Cref{fig:filter-ablation-ces}), the full \method achieves higher sPCE than \method w/o filter, with a clearer advantage when more irrelevant instances are supplied. The gap is most visible during the middle of the rollout and narrows as experimental observations accumulate. On the psychometric task (\Cref{fig:psy-aux-ablation}), the full \method maintains lower joint RMSE, with a larger and more persistent gap when more irrelevant instances are supplied. The filter can therefore provide additional protection against negative transfer.

\paragraph{Binary versus soft gating.}

\begin{figure}[!ht]
\centering
\includegraphics[width=1.0\textwidth]{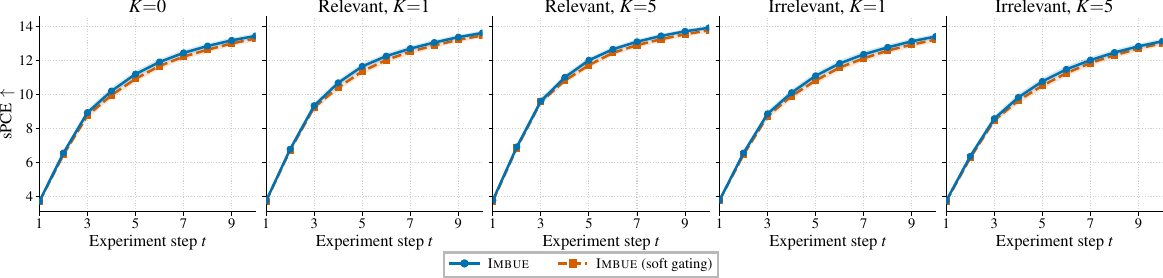}
\caption{Binary versus soft gating on CES. sPCE (mean $\pm$ 95\% CI over $2{,}000$ rollouts, $T = 10$ steps) under the five previous-instance configurations of \Cref{tab:ces-results}.}
\label{fig:ces_soft_gating}
\end{figure}

\Cref{fig:ces_soft_gating} tests whether replacing binary gating with soft gating improves sPCE on CES under the five previous-instance configurations of \Cref{tab:ces-results}. With binary gating, each in-context token is kept or excluded by thresholding $\alpha_j^{(t)}$ at $\tau=0.5$ (\Cref{sec:filter}). Soft gating instead scales each token's contribution to self-attention by its reliability score $\alpha_j^{(t)}$. The two variants achieve similar sPCE with relevant and irrelevant previous instances, indicating no clear improvement from soft gating in these experiments.

\paragraph{Sensitivity to training mixture and filter loss weight.}

\begin{figure}[!ht]
\centering
\includegraphics[width=1.0\textwidth]{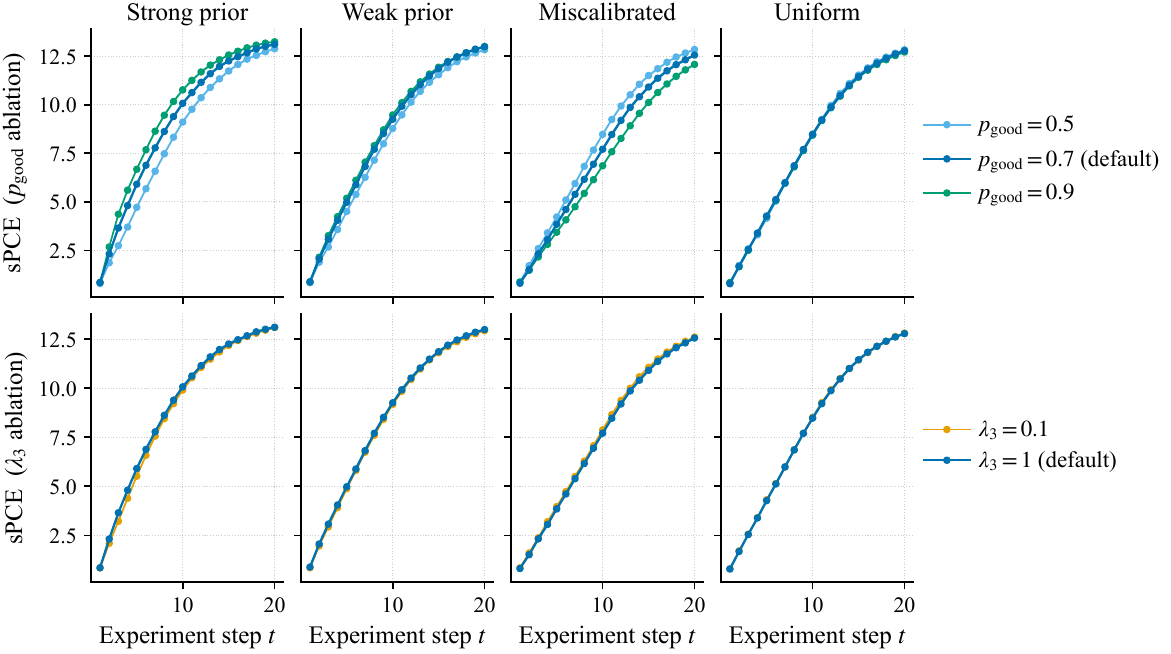}
\caption{Sensitivity sweeps on Location Finding, sPCE (mean $\pm$ 95\% CI over $2{,}000$ rollouts). \textbf{Top row}, varying the training mixture parameter $p_\text{good} \in \{0.5,\, 0.7,\, 0.9\}$, with $0.7$ as the default used in the main results. \textbf{Bottom row}, varying the filter loss weight $\lambda_3 \in \{0.1,\, 1\}$, with $1$ as the default.}

\label{fig:lf-pgood-lambda}
\end{figure}

\Cref{fig:lf-pgood-lambda} tests sensitivity to the default training settings on Location Finding. We vary $p_\text{good}$, the probability that a prior marginal is calibrated during training (Appendix~\ref{app:prior-generation}), and $\lambda_3$, the weight on the filter loss $\mathcal{L}_\text{filter}$ in \Cref{eq:total-loss}.

Increasing $p_\text{good}$ improves sPCE under strong priors but reduces it under miscalibrated priors, while weak and uniform priors are much less affected. The training mixture therefore controls a trade-off between benefiting from informative priors and robustness to misleading ones. The default $p_\text{good}=0.7$ provides a compromise within the tested range.

The default $\lambda_3=1$ and the smaller weight $\lambda_3=0.1$ give nearly identical final sPCE under all four prior conditions, so the Location Finding results are insensitive to $\lambda_3$ within the tested range.

\subsection{Deployment shifts absent from training}
\label{app:ood}
We evaluate the trained \method models under three deployment shifts not included in the training mixture (Appendix~\ref{app:prior-generation}, Appendix~\ref{app:auxiliary}). No retraining is performed, and the remaining evaluation settings follow the corresponding CES and Location Finding experiments.

\paragraph{Biased but partially relevant instances (CES).}

All $K = 5$ instances are generated as in the relevant regime, but share a systematic shift of $0.3 \times (r_l^\text{max} - r_l^\text{min})$ on a randomly selected subset of the parameters. Each instance is therefore informative about the unshifted parameters and misleading about the shifted ones. \method stays close to its $K = 0$ performance, while deployment-time concatenation degrades (\Cref{tab:ces-ablation}). Biased instances receive mean reliability scores between those of relevant and irrelevant instances (\Cref{tab:ood-alpha}). This is expected, since an instance that is right about some parameters produces weaker evidence against itself. The filter thus places a regime it has never seen between the two regimes it was trained on rather than forcing it into either.

\begin{table}[!ht]
\centering
\caption{Mean reliability score $\alpha$ over instance tokens on CES with $K = 5$ instances (mean $\pm$ 95\% CI, $2{,}000$ rollouts).}
\label{tab:ood-alpha}
\setlength{\tabcolsep}{8pt}
\begin{tabular}{l c c c}
\toprule
Step $t$ & Relevant & Biased (shift) & Irrelevant \\
\midrule
$2$  & $0.818 \pm 0.004$ & $0.677 \pm 0.012$ & $0.479 \pm 0.010$ \\
$10$ & $0.906 \pm 0.002$ & $0.621 \pm 0.016$ & $0.246 \pm 0.008$ \\
\bottomrule
\end{tabular}
\end{table}

\paragraph{A different noise model in the previous instances (CES).}

The instance parameters are generated as in the relevant regime, but the instance outcomes carry heavier-tailed Student-$t$ noise at a larger scale in place of the Gaussian noise used during training. The live experiment is unchanged and uses the training likelihood. Under the oracle definition of \Cref{eq:label-aux}, relevant instances remain reliable because their parameters stay close to $\theta^\star$. \Cref{tab:ood-noise} shows that relevant instances remain beneficial, while irrelevant instances leave performance close to the $K = 0$ condition.

\begin{table}[!ht]
\centering
\caption{CES with $K = 5$ instances whose outcomes carry Student-$t$ noise ($\nu = 3$, scale $3\sigma_{\mathrm{CES}}$) instead of the Gaussian noise used in training. sPCE at $T = 10$ (mean $\pm$ 95\% CI, $2{,}000$ rollouts).}
\label{tab:ood-noise}
\setlength{\tabcolsep}{8pt}
\begin{tabular}{l c}
\toprule
Condition & sPCE \\
\midrule
$K = 0$ & $13.46 \pm 0.15$ \\
$K = 5$ relevant, Gaussian noise (in-dist.) & $13.93 \pm 0.12$ \\
$K = 5$ relevant, Student-$t$ noise (shift) & $14.00 \pm 0.12$ \\
$K = 5$ irrelevant, Gaussian noise (in-dist.) & $13.14 \pm 0.16$ \\
$K = 5$ irrelevant, Student-$t$ noise (shift) & $13.32 \pm 0.16$ \\
\bottomrule
\end{tabular}
\end{table}

\paragraph{Priors with incorrect support (Location Finding).}

We test a more severe form of prior misspecification in which the supplied prior assigns exactly zero probability to the bin containing the true parameter. We first draw $\theta^\star\sim\mathcal U[0,1]^4$, and only then construct the prior histogram so that it excludes the corresponding bin. We consider two variants: a \emph{truncated} prior that concentrates its mass in a window away from $\theta^\star$, and a \emph{boundary} prior that places its mass at the two ends of the support while $\theta^\star$ lies in the interior. This differs fundamentally from training, where $\theta^\star$ is sampled from the supplied histogram and therefore always lies within its support. We observe no sharp degradation in sPCE under either form of prior misspecification (\Cref{tab:ood-prior}).

\begin{table}[!ht]
\centering
\caption{Location Finding under priors that assign zero mass to the bin containing $\theta^\star$. sPCE (mean $\pm$ 95\% CI, $2{,}000$ rollouts) at four experiment steps, alongside the three in-distribution prior types.}
\label{tab:ood-prior}
\setlength{\tabcolsep}{8pt}
\begin{tabular}{l c c c c}
\toprule
Prior & $t = 2$ & $t = 5$ & $t = 11$ & $t = 20$ \\
\midrule
Strong (in-dist.)        & $2.26 \pm 0.07$ & $5.77 \pm 0.11$ & $10.49 \pm 0.11$ & $13.03 \pm 0.06$ \\
Uniform (in-dist.)       & $1.72 \pm 0.06$ & $4.25 \pm 0.09$ & $9.10 \pm 0.12$  & $12.83 \pm 0.07$ \\
Miscalibrated (in-dist.) & $1.56 \pm 0.05$ & $3.90 \pm 0.08$ & $8.43 \pm 0.11$  & $12.61 \pm 0.07$ \\
\midrule
Truncated (shift)        & $1.46 \pm 0.05$ & $3.89 \pm 0.08$ & $8.54 \pm 0.11$  & $12.56 \pm 0.07$ \\
Boundary (shift)         & $1.76 \pm 0.06$ & $4.27 \pm 0.09$ & $9.12 \pm 0.11$  & $12.77 \pm 0.06$ \\
\bottomrule
\end{tabular}
\end{table}

Taken together, these experiments probe three distinct deployment shifts absent from training: biased but partially relevant previous instances, a different noise model in the previous instances, and priors assigning zero mass to the truth. \method does not degrade sharply in any of them. They are three specific, plausible shifts and do not establish robustness to arbitrary deployment mismatch.

\subsection{Sensitivity to the candidate design pool}
\label{app:pool-sensitivity}

We keep the trained policies fixed and vary the candidate pool size at deployment from $N_q^\text{eval} = 200$ to $8{,}000$, using i.i.d.\ uniform or Sobol pools on CES (\Cref{tab:pool-ces}) and Location Finding (\Cref{tab:pool-lf}). The main-text setting is $N_q^\text{eval} = 2{,}000$ with uniform pools.

\Cref{tab:pool-ces,tab:pool-lf} show that the main qualitative conclusions hold across the tested pool sizes and constructions: relevant instances remain beneficial, irrelevant instances leave performance close to the $K=0$ condition, and the ordering of the prior conditions is preserved. Gains from enlarging the pool diminish beyond the default size. Larger pools give denser coverage of the design space and are likely to become more useful as the design dimension grows. In very high-dimensional design spaces, enumeration is not realistic, but since the design head produces a differentiable pre-normalization score for each candidate, this score could in principle be optimized directly with respect to the design in continuous space.

\begin{table}[!ht]
\centering
\small
\caption{CES. sPCE at $T = 10$ as a function of the candidate design pool $N_q^\text{eval}$ and construction, with the trained policy fixed ($2{,}000$ rollouts; 95\% CIs are approximately $\pm 0.14$). The main-text setting is $N_q^\text{eval} = 2{,}000$ with uniform pools.}
\label{tab:pool-ces}
\setlength{\tabcolsep}{6pt}
\begin{tabular}{l l c c c c c c}
\toprule
Condition & Pool & $N_q^\text{eval} = 200$ & $500$ & $1{,}000$ & $2{,}000$ & $4{,}000$ & $8{,}000$ \\
\midrule
$K = 0$          & Uniform & $13.03$ & $13.23$ & $13.40$ & $13.46$ & $13.54$ & $13.58$ \\
                 & Sobol   & $13.15$ & $13.35$ & $13.48$ & $13.58$ & $13.69$ & $13.70$ \\
\midrule
$K = 1$ rel.     & Uniform & $13.25$ & $13.48$ & $13.61$ & $13.63$ & $13.75$ & $13.81$ \\
                 & Sobol   & $13.47$ & $13.62$ & $13.74$ & $13.84$ & $13.90$ & $13.96$ \\
\midrule
$K = 5$ rel.     & Uniform & $13.60$ & $13.79$ & $13.90$ & $13.93$ & $13.98$ & $14.02$ \\
                 & Sobol   & $13.59$ & $13.73$ & $13.90$ & $13.93$ & $13.98$ & $14.00$ \\
\midrule
$K = 1$ irrel.   & Uniform & $12.92$ & $13.19$ & $13.33$ & $13.42$ & $13.52$ & $13.55$ \\
                 & Sobol   & $12.83$ & $13.10$ & $13.26$ & $13.34$ & $13.45$ & $13.48$ \\
\midrule
$K = 5$ irrel.   & Uniform & $12.83$ & $13.07$ & $13.20$ & $13.14$ & $13.34$ & $13.37$ \\
                 & Sobol   & $12.90$ & $13.13$ & $13.24$ & $13.37$ & $13.47$ & $13.54$ \\
\bottomrule
\end{tabular}
\end{table}

\begin{table}[!ht]
\centering
\small
\caption{Location Finding. sPCE at $T = 20$ as a function of the candidate design pool $N_q^\text{eval}$ and construction, with the trained policy fixed ($2{,}000$ rollouts; 95\% CIs are approximately $\pm 0.06$). The main-text setting is $N_q^\text{eval} = 2{,}000$ with uniform pools.}
\label{tab:pool-lf}
\setlength{\tabcolsep}{6pt}
\begin{tabular}{l l c c c c c c}
\toprule
Prior & Pool & $N_q^\text{eval} = 200$ & $500$ & $1{,}000$ & $2{,}000$ & $4{,}000$ & $8{,}000$ \\
\midrule
Strong        & Uniform & $11.91$ & $12.81$ & $13.05$ & $13.03$ & $13.13$ & $13.07$ \\
              & Sobol   & $12.19$ & $12.93$ & $13.12$ & $13.13$ & $13.10$ & $13.09$ \\
\midrule
Weak          & Uniform & $11.88$ & $12.64$ & $12.84$ & $12.89$ & $12.99$ & $12.94$ \\
              & Sobol   & $12.07$ & $12.81$ & $12.95$ & $12.94$ & $12.98$ & $12.95$ \\
\midrule
Uniform       & Uniform & $11.79$ & $12.50$ & $12.69$ & $12.83$ & $12.80$ & $12.79$ \\
              & Sobol   & $11.99$ & $12.64$ & $12.71$ & $12.80$ & $12.78$ & $12.79$ \\
\midrule
Miscalibrated & Uniform & $11.67$ & $12.31$ & $12.49$ & $12.61$ & $12.55$ & $12.55$ \\
              & Sobol   & $11.82$ & $12.44$ & $12.55$ & $12.65$ & $12.61$ & $12.57$ \\
\bottomrule
\end{tabular}
\end{table}

\subsection{Computational overhead of in-context conditioning}
\label{app:overhead}

\Cref{tab:gp-time} compares \method with GP baselines on the active learning task. Here we break the overhead of in-context conditioning down by module on CES and Location Finding, separating the in-context token pathway from the reliability filter.

\textit{Parameters.} With a fixed filter architecture and embedding dimension ($D=32$), $f_\omega$ (\Cref{eq:gate-logit}) has $9{,}089$ parameters, independent of $K$, the number of histogram bins $N_B$, and the task. In our experimental configurations, this amounts to $6.3\%$ of the full CES model ($145{,}408$ parameters) and $7.3\%$ of the full Location Finding model ($123{,}744$ parameters). Beyond the filter, the additional conditioning parameters are primarily in the per-instance encoder $g_\text{aux}$ for CES ($38{,}112$ parameters, plus $64$ summary-token parameters), and in the histogram prior embedder for Location Finding ($17{,}056$ parameters at $N_B=100$).

\textit{Per-step latency.} \Cref{tab:overhead-latency} shows that on CES, encoding previous instances increases inference time, and enabling the reliability filter adds a further cost. On Location Finding, the comparison isolates the additional time required to filter prior tokens. For both tasks, the full model keeps per-step inference time in the millisecond range. The $N_q^\text{eval} = 2{,}000$ candidate tokens dominate the sequence, so shortening the in-context segment from $K \cdot T = 50$ concatenated observations to $K = 5$ instance tokens brings little latency benefit at this scale. The advantage of the token interface appears at larger $K$ and $T$, since the in-context segment of ALINE\textsuperscript{*} grows as $K \cdot T$ whereas that of \method grows as $K$, so concatenation eventually exceeds the sequence-length budget.

\begin{table}[!ht]
\centering
\caption{Per-step inference latency of \method by module (ms per step, mean $\pm$ standard deviation over $10$ runs; NVIDIA H100, batch size $1$, $N_q^\text{eval} = 2{,}000$). On CES the in-context input is $K = 5$ instances; on Location Finding it is the prior histogram.}
\label{tab:overhead-latency}
\setlength{\tabcolsep}{8pt}
\begin{tabular}{l c c}
\toprule
Configuration & CES & Location Finding \\
\midrule
No in-context tokens      & $2.55 \pm 0.003$ & --- \\
In-context tokens, w/o filter      & $3.62 \pm 0.003$ & $2.56 \pm 0.002$ \\
In-context tokens + filter (full)  & $4.57 \pm 0.005$ & $3.28 \pm 0.003$ \\
\bottomrule
\end{tabular}
\end{table}

\textit{Accuracy--efficiency trade-off.} 
On CES, the instance-token interface improves sPCE over ALINE-aux under relevant, biased, and irrelevant inputs. Adding the reliability filter further improves mean sPCE under biased and irrelevant inputs while maintaining comparable performance under relevant input. On Location Finding, the filter has little effect on sPCE under the tested prior conditions, but provides interpretable assessments of prior reliability.

\section{Computational resources and software}
\label{app:computation_resources}

All experiments, including model development, hyperparameter tuning, baseline runs, and preliminary analyses, were conducted on NVIDIA H100 80GB GPUs. The total compute consumed across all stages is approximately $4{,}000$ GPU hours. Training an \method model for $10^5$ epochs takes roughly $7$ hours per benchmark. The core code base is implemented in PyTorch~\citep{paszke2019pytorch} (\url{https://pytorch.org/}, License: modified BSD license). The GP baselines use scikit-learn~\citep{pedregosa2011scikit} (\url{https://scikit-learn.org/}, License: modified BSD license). For DAD~\citep{foster2021deep} and RL-BOED~\citep{blau2022optimizing}, we use the PyTorch implementations at \url{https://github.com/e-wxy/DAD} (License: MIT) and \url{https://github.com/huangdaolang/RL-BOED}, respectively. MT-GP~\citep{bonilla2007multi} is our own PyTorch implementation described in Appendix~\ref{app:exp-gp}. For ALINE~\citep{huang2025aline}, we use the official implementation (\url{https://github.com/huangdaolang/ALINE}, License: MIT), on which our code base also builds. We use the \texttt{questplus} package (\url{https://github.com/hoechenberger/questplus}, License: GPL-3.0) for QUEST+ and \texttt{Psi-staircase} (\url{https://github.com/NNiehof/Psi-staircase}, License: GPL-3.0) for Psi-marginal.

\clearpage

%%%%%%%%%%%%%%%%%%%%%%%%%%%%%%%%%%%%%%%%%%%%%%%%%%%%%%%%%%%%

\newpage

\end{document}